\documentclass{article}
\usepackage{iclr2027_conference,times}

\iclrfinalcopy

\usepackage{amsmath,amsfonts,bm}

\def\eqref#1{equation~\ref{#1}}

\def\1{\bm{1}}

\DeclareMathAlphabet{\mathsfit}{\encodingdefault}{\sfdefault}{m}{sl}
\SetMathAlphabet{\mathsfit}{bold}{\encodingdefault}{\sfdefault}{bx}{n}

\usepackage{hyperref}
\usepackage{url}

\usepackage{graphicx}
\usepackage{booktabs}
\usepackage{multirow}
\usepackage{amsmath}
\usepackage{amssymb}
\usepackage{xcolor}
\usepackage{wrapfig}
\usepackage{listings}
\usepackage{array}

\definecolor{pyKeyword}{HTML}{0000CC}
\definecolor{pyBuiltin}{HTML}{900090}
\definecolor{pyString}{HTML}{BA2121}
\definecolor{pyComment}{HTML}{408080}
\definecolor{pyNumber}{HTML}{008000}
\definecolor{pyBg}{HTML}{F7F7F9}
\definecolor{pyRule}{HTML}{D0D0D8}
\lstdefinestyle{dattripy}{
  language=Python,
  basicstyle=\ttfamily\scriptsize,
  keywordstyle=\color{pyKeyword}\bfseries,
  stringstyle=\color{pyString},
  commentstyle=\color{pyComment}\itshape,
  emph={DataSelectionCallback,HookManager,HookManagerConfig},
  emphstyle=\color{pyBuiltin}\bfseries,
  emph={[2]True,False,None},
  emphstyle={[2]\color{pyKeyword}},
  literate=*{0.5}{{\textcolor{pyNumber}{0.5}}}{3},
  backgroundcolor=\color{pyBg},
  frame=single, rulecolor=\color{pyRule}, framerule=0.4pt,
  xleftmargin=6pt, xrightmargin=6pt, framexleftmargin=6pt, framexrightmargin=6pt,
  aboveskip=4pt, belowskip=2pt,
  columns=fullflexible, keepspaces=true, breaklines=true,
  showstringspaces=false,
  tabsize=4,
}

\usepackage{pifont}
\newcommand{\cmark}{\ding{51}}
\newcommand{\xmark}{\ding{55}}
\newcommand{\pmark}{$\boldsymbol{\sim}$}

\providecommand{\dattri}{\texttt{dattri-LLM}}
\newcommand{\code}[1]{\texttt{#1}}
\newcommand{\oom}{\textsc{oom}}

\title{{dattri-LLM}: A Unified and Efficient Library for Training Data Attribution at LLM Scale}

\author{
\textbf{Shixuan Liu}$^1$, \textbf{Tongli Zhou}$^2$, \textbf{Junwei Deng}$^1$, \textbf{Pingbang Hu}$^1$, \textbf{Jiaqi W. Ma}$^1$ \\
\normalfont $^{1}$University of Illinois at Urbana-Champaign \quad
\normalfont $^{2}$Google
}

\begin{document}

\maketitle
\lhead{Preprint}

\begin{abstract}
Training data attribution (TDA) estimates the contribution of individual training examples to model outputs. Most scalable TDA methods rely on per-example gradients, whose computation and use at LLM scale pose challenges in efficiency, compatibility, and extensibility. We introduce \dattri{}, a TDA library that makes gradient-based attribution more practical at scale. For efficiency, \dattri{} uses compact gradient representations and dynamically routes gradient operations based on a cost model. For compatibility, its capture mechanism collects per-example gradients from existing training loops that call \texttt{backward()}, without requiring changes to the loop or its configuration. This includes distributed training with DDP and FSDP and pipelines built with HuggingFace Transformers, TRL, and OLMo. For extensibility, \dattri{} exposes reusable gradient operations and training-time callbacks for implementing attribution methods and applications. These interfaces support a variety of attribution methods, including gradient similarity, curvature-based influence, and trajectory-based methods, as well as applications that act on gradients during training, such as online data selection. On the same hardware and workload, \dattri{} achieves 3.2$\times$ the throughput of the fastest competing library on average, scales multiple attribution methods to $110$B-parameter models across four H200 GPUs, and offers superior attribution fidelity--cost trade-offs across a range of models with different model families and scales. The source code of \dattri{} is available at \url{https://github.com/TRAIS-Lab/dattri-llm}.
\end{abstract}

\section{Introduction}
\label{sec:intro}

Training data attribution (TDA)~\citep{deng2025survey} estimates the contribution of individual training examples to a model's outputs, with applications in model debugging~\citep{koh2017understanding,yeh2018representer,kwon2024datainf}, data selection~\citep{xia2024less,wang2024greats,hu2026dr}, and data valuation~\citep{ghorbani2019data,jia2019towards,kwon2021beta}. Many scalable TDA methods compare gradients of a query objective with per-example training gradients, either directly, as in TracIn~\citep{pruthi2020estimating}, or after transformations such as projection and preconditioning, as in TRAK and EK-FAC-based influence estimation~\citep{park2023trak,grosse2023studying}. Although their formulations differ, these methods share computational building blocks: obtaining per-example gradients and applying operations such as projection, preconditioning, and inner products.

At LLM scale, these computations pose three systems challenges. The first is \emph{efficiency}. A single per-example gradient can contain billions of entries, making its explicit construction and storage expensive~\citep{schioppa2022scaling,grosse2023studying,park2023trak}. Structured representations can reduce this cost~\citep{li2021large}, but their relative efficiency depends on the layer shape and the operation being performed. Efficient execution therefore requires choosing suitable representations for different gradient operations. The second is \emph{compatibility}. Existing libraries~\citep{deng2024dattri,quirke2026bergson} offer task-based attribution workflows, trainer integrations, and custom-loop gradient collectors. However, using these capabilities within an existing training pipeline can still require changes to the training code or framework-specific integration. The third is \emph{extensibility}. Implementing an attribution method requires reusable components, such as gradient capture, projection, and curvature estimation. Applications that act during training additionally need defined callback points at which they can access captured gradients and inspect or modify training state. Supporting both requires interfaces that separate method-specific logic from more general modules, such as gradient capture and execution.

To close these gaps, we introduce \dattri{}, a library for \textbf{Da}ta \textbf{Attri}bution at \textbf{LLM} scale, built around a unified gradient interface. Its design addresses these three challenges:

\paragraph{Efficiency: exact gradient representations with FLOP-aware routing.}
\dattri{} supports two exact representations of per-example gradients for supported layers: \emph{factorized} and \emph{materialized}. A FLOP-aware cost model selects between them for each layer and operation, reducing execution cost without introducing additional approximation into the attribution computation. Shared gradient operations make this routing available across attribution methods (\S\ref{sec:efficiency}). Across the methods and projection regimes evaluated in \S\ref{sec:scaling}, \dattri{} achieves an average speedup of $3.2\times$ over the fastest competing library for each configuration.

\paragraph{Compatibility: non-invasive gradient capture from existing training loops.}
Alongside an end-to-end attribution interface, \dattri{} provides an \emph{open-loop} interface that uses autograd hooks to capture per-example gradients during backward passes in existing training loops. Users enable capture by enclosing the training call in a context manager, without editing the training loop, optimizer, or trainer. We demonstrate this integration with HuggingFace \code{Trainer}~\citep{wolf2020transformers}, TRL~\citep{vonwerra2020trl}, and OLMo~\citep{groeneveld2024olmo}, and evaluate capture under DDP~\citep{li2020pytorch} and FSDP~\citep{zhao2023pytorch} (\S\ref{sec:compat}). Captured gradients can also be cached and reused across queries and attribution methods when the cache satisfies their gradient and checkpoint requirements, avoiding repeated training-side computation.

\paragraph{Extensibility: interfaces for new methods and training-time applications.}
\dattri{} exposes an \code{Attributor} interface for implementing gradient-based attribution methods using a common gradient stream and shared gradient operations. Its \code{HookManagerCallback} interface provides callback points during gradient capture, allowing applications to inspect or modify training state using the captured gradients (\S\ref{sec:extend}). We demonstrate this capability by integrating a recent online data-selection method into an existing fine-tuning loop through a callback (\S\ref{sec:dataselect}).

\begin{figure}[t]
  \centering
  \includegraphics[width=\linewidth]{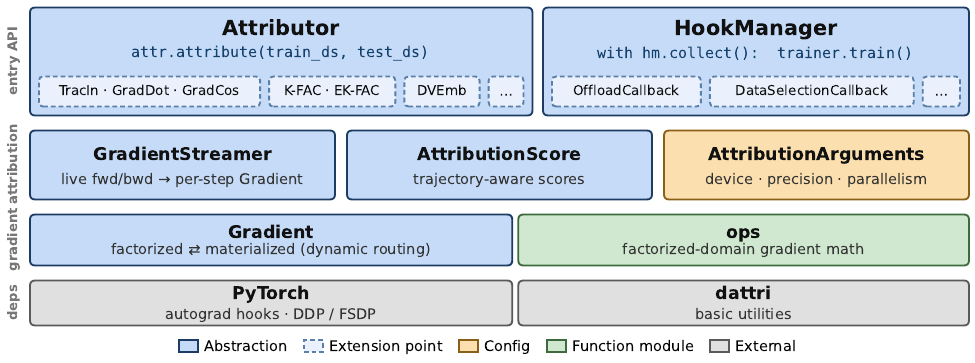}
  \caption{\textbf{Architecture overview of \dattri{}.} From bottom to top, the stack comprises external dependencies (gray), per-example gradient representations and operations (green), shared attribution components, and the entry API. Solid blue boxes denote classes that users instantiate and call; orange denotes configuration. Dashed boxes denote abstract base classes that serve as extension points: subclassing \code{Attributor} adds an attribution method, while subclassing \code{HookManagerCallback} adds an application that acts during gradient capture. }
  \label{fig:arch}
  \vspace{-10pt}
\end{figure}

\section{Related Work}
\label{sec:related}

\begin{table}[t]
  \centering
  \small
  \setlength{\tabcolsep}{5pt}
  \caption{Comparison of recent libraries for LLM-scale TDA. \cmark{}/\xmark{} indicates supported / not supported; \pmark{} indicates partial support: Kronfluence uses a cost model, but it always builds full gradients on the query side and optionally builds training-side gradients on demand. Bergson integrates with a wrapped HuggingFace \code{Trainer} through callbacks that record gradients without acting on the training state. LogIX supports DDP but not FSDP. Kronfluence and Bergson provide interfaces for customizing specific steps of attribution, such as gradient collection or preconditioning, but not for defining a new end-to-end attribution method. }
  \label{tab:libraries}
  \begin{tabular}{lcccc}
    \toprule
    & \dattri{} & LogIX & Kronfluence & Bergson \\
    \midrule
    \multicolumn{5}{l}{\emph{Efficiency}} \\
    Dimension reduction
      & \cmark & \cmark & \xmark & \cmark \\
    Training-gradient reuse
      & \cmark & \cmark & \xmark & \cmark \\
    Cost-based operation routing
      & \cmark & \xmark & \cmark & \xmark \\
   Full gradients built on demand
      & \cmark & \xmark & \pmark & \xmark \\
    \midrule
    \multicolumn{5}{l}{\emph{Compatibility}} \\
    Integration with existing training
      & \cmark & \xmark & \xmark & \pmark \\
    Distributed execution
      & \cmark & \pmark & \cmark & \cmark \\
    \midrule
    \multicolumn{5}{l}{\emph{Extensibility}} \\
    Interface for new attribution methods
      & \cmark & \xmark & \pmark & \pmark \\
    Interface for training-time applications
      & \cmark & \xmark & \xmark & \xmark \\
    \bottomrule
  \end{tabular}
  \vspace{-10pt}
\end{table}

We review TDA methods, whose shared building blocks motivate the design of \dattri{}, and existing TDA libraries, against which we position this library.

\paragraph{TDA methods.}
TDA methods estimate how individual training examples contribute to a model's behavior, and are commonly grouped by the principles underlying their influence measures. \emph{Influence-function-based} methods~\citep{koh2017understanding} use local approximations to estimate the effect of removing training examples. Scalable variants use Arnoldi iteration~\citep{schioppa2022scaling}, Kronecker-factored curvature (K-FAC and EK-FAC)~\citep{martens2015optimizing,george2018fast,grosse2023studying}, or closed-form inverse-curvature approximations such as DataInf~\citep{kwon2024datainf}. TRAK~\citep{park2023trak} and LoGRA~\citep{choe2026your} reduce computational costs through gradient projection. \emph{Weighted marginal contribution} methods, such as Data Shapley~\citep{ghorbani2019data}, average an example's marginal contribution across training subsets under a specified weighting. \emph{Training-dynamics-based} methods use information along the optimization trajectory. TracIn~\citep{pruthi2020estimating} accumulates gradient inner products over checkpoints, while Grad-Dot and Grad-Cos~\citep{charpiat2019input} evaluate gradient similarity at a single checkpoint. In-Run Data Shapley~\citep{wang2024inrun} assigns contributions within a realized training run, while data value embeddings~\citep{wang2025capturing} and SOURCE~\citep{bae2024source} approximate how an example's influence propagates through subsequent training updates. AdamW-influence~\citep{deng2026faithfultrajectorybaseddataattribution} extends trajectory-based influence estimation to account for AdamW's first- and second-moment states, adaptive scaling, and weight decay. \emph{Simulator-based} methods, including Datamodels~\citep{ilyas2022datamodels} and Simfluence~\citep{guu2023simfluence}, fit surrogate models to observations from training runs to predict model behavior under different training subsets or curricula. Other approaches include representer points~\citep{yeh2018representer}, which decompose predictions using weighted similarities between learned features. Across these families, gradient-based estimators support applications such as data selection with LESS~\citep{xia2024less} and pretraining attribution with TrackStar~\citep{chang2024trackstar}, while techniques such as gradient sparsification in GraSS~\citep{hu2026grass} improve scalability. Despite their differences, these methods share the same building blocks of per-example gradients, projection or preconditioning, and pairwise scoring, which is the layer at which \dattri{} is organized. We refer readers to \citet{deng2025survey} for a comprehensive review.

\paragraph{TDA libraries.}
General-purpose TDA libraries such as \code{dattri}~\citep{deng2024dattri}, \code{pyDVL}~\citep{transferlab2024pydvl}, OpenDataVal~\citep{jiang2023opendataval}, and Influenciae~\citep{picard2024influenciae} support data valuation and attribution, including game-theoretic and influence-function estimators. However, most are designed for smaller machine learning models that predate modern LLMs. Among them, \code{dattri} provides projected attributors that scale to billion-parameter models on a single GPU, but does not support attribution for models with tens or hundreds of billions of parameters across multiple GPUs. Libraries specifically developed for LLM-scale attribution include Kronfluence~\citep{grosse2023studying}, which implements K-FAC and EK-FAC influence functions with distributed and token-level scoring; LogIX~\citep{choe2026your}, which captures projected gradients through patched HuggingFace or lightning trainers; and Bergson~\citep{quirke2026bergson}, which supports collecting projected or full gradients into a store and integrates with the HuggingFace \code{Trainer}. Table~\ref{tab:libraries} compares these libraries with \dattri{} along the three design principles of \S\ref{sec:design} in terms of efficiency-related capabilities, compatibility with existing training pipelines, and interfaces for extension. Across these rows, \dattri{} covers the efficiency, compatibility, and extensibility requirements together.

\section{Design}
\label{sec:design}

In this section we introduce the design principles of \dattri{}. We first give an overview of its architecture (\S\ref{sec:architecture}), and then describe how each of the three properties from \S\ref{sec:intro} is realized: efficiency through the choice of gradient representation (\S\ref{sec:efficiency}), compatibility through non-invasive gradient capture (\S\ref{sec:compat}), and extensibility through two designated extension points (\S\ref{sec:extend}). Implementation details are given in Appendix~\ref{app:design}.

\subsection{Architecture overview}
\label{sec:architecture}

\dattri{} is organized into four tiers (Figure~\ref{fig:arch}). The bottom tier holds the dependencies: PyTorch~\citep{paszke2019pytorch}, whose autograd hooks and DDP/FSDP wrappers the library builds on, and \code{dattri}~\citep{deng2024dattri}, from which it imports basic utilities such as random projectors. The \emph{gradient} tier defines the \code{Gradient} object, which stores a batch of per-example gradients across selected layers, using either \emph{factorized} or \emph{materialized} representation that we will introduce in \S\ref{sec:efficiency}, together with the mathematical operations on it. The \emph{attribution} tier holds the logic shared by all attribution methods: the \code{GradientStreamer} that produces per-step gradients from a model, the \code{AttributionScore} result container, the \code{AttributionArguments} configuration, and the callbacks that act on the capture loop. The top tier is the entry API, which offers two ways in. \code{Attributor} is the \emph{closed-loop} interface: given a model, training set, and query set, it directly returns scores for pre-built attribution methods, including TracIn~\citep{pruthi2020estimating}, LESS~\citep{xia2024less}, K-FAC and EK-FAC influence functions~\citep{martens2015optimizing,george2018fast,grosse2023studying}, data-value embeddings~\citep{wang2025capturing} and AdamW-Influence~\citep{deng2026faithfultrajectorybaseddataattribution}. \code{HookManager} is the \emph{open-loop} interface: it captures per-example gradients from \code{.backward()} executed inside a \code{with} block, so attribution can be attached to a training loop the user already owns (\S\ref{sec:compat}).

\subsection{Efficiency: exact gradient representations with FLOP-aware routing}
\label{sec:efficiency}

Computing and comparing per-example gradients is a major cost of TDA at LLM scale. \dattri{} exploits the fact that the same gradient admits two exact representations whose computational costs differ across operations and layer shapes. It selects the representation separately for each layer and operation using a FLOP-aware cost model.

\paragraph{Two representations of gradients.}
\label{sec:repr}
Consider a linear layer $z=Wa$ with $W\in\mathbb{R}^{N_o\times N_i}$, per-example input $a\in\mathbb{R}^{N_i\times T}$ over $T$ token positions, and output $z\in\mathbb{R}^{N_o\times T}$. Let $\mathrm{D}$ denote the differentiation operator $\partial\ell/\partial(\cdot)$ and $g=\mathrm{D} z\in\mathbb{R}^{N_o\times T}$ denote the gradient of the per-example loss $\ell$ with respect to the layer output. By the chain rule, the weight gradient is $\mathrm{D}W=\partial\ell/\partial W=g\,a^\top$, or, in flattened form,
\begin{equation}
    \gamma:=\operatorname{vec}(\mathrm{D}W)
    =\sum_{t=1}^{T} a_t\otimes g_t,
    \label{eq:canonical-factorization}
\end{equation}
where $a_t$ and $g_t$ are the columns at token position $t$, $\operatorname{vec}(\cdot)$ stacks matrix columns, and $\otimes$ denotes the Kronecker product. We call the dense matrix $\mathrm{D}W$ the \emph{materialized} representation and the pair $(a,g)$ the \emph{factorized} representation. The materialized gradient can be recovered exactly from the factors. The factorized form stores $T(N_i+N_o)$ values, compared with $N_iN_o$ for the materialized form. It therefore uses less storage when $T<N_iN_o/(N_i+N_o)$, favoring wide layers and short sequences. Convolutional and embedding layers also admit related factorizations (Appendix~\ref{app:layers}), extending this representation beyond linear layers.

\paragraph{Gradient postprocessing.}
\label{sec:postproc}
\dattri{} supports optional transforms before scoring, including \emph{dimension reduction} and \emph{optimizer preconditioning}. Dimension reduction can operate on either the factors or the materialized gradient. \emph{Factor projection} applies random matrices $P_a\in\mathbb{R}^{k_a\times N_i}$ and $P_g\in\mathbb{R}^{k_g\times N_o}$ to the two factors:
\begin{equation}
    \tilde a_t=P_a a_t,\qquad
    \tilde g_t=P_g g_t,\qquad
    \sum_{t=1}^{T}\tilde a_t\otimes\tilde g_t
    =(P_a\otimes P_g)\gamma,
    \label{eq:factor-projection}
\end{equation}
where $k_a$ and $k_g$ are the projected dimensions. This preserves the factorized structure in Eq.~\ref{eq:canonical-factorization}, allowing subsequent operations to use the same formulas in the projected space~\citep{choe2026your,hu2026grass}. \emph{Materialized projection} instead applies a single random matrix $P\in\mathbb{R}^{k_{\text{dense}}\times(N_iN_o)}$ to the flattened gradient, yielding $\tilde\gamma=P\gamma$, where $k_{\text{dense}}$ is the projection dimension for the dense gradient. This approach, used in TRAK~\citep{park2023trak}, also applies to layers without a factorized gradient representation. Supported random projections include Gaussian~\citep{johnson1984extensions}, Rademacher~\citep{achlioptas2003database}, sparse Johnson--Lindenstrauss transforms~\citep{kane2014sparser}, and the sparsified projection of GraSS~\citep{hu2026grass}. Alternatively, a random mask selects a subset of gradient entries, which can be computed directly from the factors without materializing the full gradient (Appendix~\ref{app:projection}).

\emph{Optimizer preconditioning} transforms a per-example gradient into an update direction using the training optimizer's current state. For Adam~\citep{kingma2015adam}, the direction for sample $i$ is
\begin{equation}
    \Gamma^{(i)}
    =\frac{\hat m^{(i)}}{\sqrt{\hat v^{(i)}}+\epsilon},
    \qquad
    m^{(i)}=\beta_1 m+(1-\beta_1)\gamma^{(i)},
    \qquad
    v^{(i)}=\beta_2 v+(1-\beta_2)(\gamma^{(i)})^2,
    \label{eq:adam-direction}
\end{equation}
where parenthesized superscripts index samples, $(m,v)$ are the optimizer's moment estimates for the corresponding parameters before the current step, $\beta_1,\beta_2$ are their decay rates, and $\epsilon>0$ ensures numerical stability. At optimizer step $\tau$, the bias-corrected updated moments are $\hat m^{(i)}=m^{(i)}/(1-\beta_1^\tau)$ and $\hat v^{(i)}=v^{(i)}/(1-\beta_2^\tau)$ (Appendix~\ref{app:optcapture}). This transform enables optimizer-aware methods such as LESS~\citep{xia2024less} and AdamW-influence~\citep{deng2026faithfultrajectorybaseddataattribution}.

\paragraph{Factorized inner product.}
For samples $i$ and $j$ with $T_i$ and $T_j$ token positions, respectively, the Kronecker-product inner-product identity gives
\begin{equation}
    \left\langle\gamma^{(i)},\gamma^{(j)}\right\rangle
    =\left\langle\mathrm{D}W^{(i)},\mathrm{D}W^{(j)}\right\rangle_F
    =\sum_{t=1}^{T_i}\sum_{s=1}^{T_j}
    \left\langle a_t^{(i)},a_s^{(j)}\right\rangle
    \left\langle g_t^{(i)},g_s^{(j)}\right\rangle,
    \label{eq:ghostdot}
\end{equation}
where $\langle\cdot,\cdot\rangle_F$ denotes the Frobenius inner product. Thus, exact gradient inner products can be computed directly from the factors without materializing the weight gradients. Factorized representations also support the preconditioned gradient comparisons used in K-FAC~\citep{martens2015optimizing} and EK-FAC~\citep{george2018fast}. Retaining the token axis additionally enables attribution at token granularity.

\paragraph{FLOP-aware routing.}
\label{sec:routing}
Eq.~\ref{eq:ghostdot} compares all pairs of token positions, so its cost is quadratic in sequence length. Alternatively, materializing the gradients incurs a construction cost that can be amortized over subsequent dense inner products. For a cross-Gram matrix between batches of $B_1$ and $B_2$ examples, each with $T$ token positions, the factorized and materialized routes cost $O\!\left(B_1B_2T^2(N_i+N_o)\right)$ and $O\!\left((B_1+B_2)TN_iN_o+B_1B_2N_iN_o\right)$, respectively. Neither route is uniformly cheaper: short sequences and wide layers favor factorized computation, while long sequences and large comparison batches favor materialization. Factor projection replaces $N_i,N_o$ with $k_a,k_g$ in these costs and can change which route is preferable. \dattri{} therefore estimates the FLOP cost of each route per layer and operation and selects the cheaper one (Appendix~\ref{app:routing}). An ablation in Appendix~\ref{app:routingexp} evaluates the benefit of this routing strategy.

\subsection{Compatibility: non-invasive gradient capture from arbitrary training loops}
\label{sec:compat}

In \dattri{}, attribution can be integrated into an existing pipeline without modifying its training loop. Specifically, the \emph{open-loop} interface \code{HookManager} registers autograd hooks on selected layers and captures the two factors in Eq.~\ref{eq:canonical-factorization} during each backward pass triggered by \code{.backward()}. This mechanism applies uniformly to pretraining, supervised fine-tuning, and reinforcement-learning loops. For example, unmodified HuggingFace \code{Trainer}~\citep{wolf2020transformers}, TRL~\citep{vonwerra2020trl}, and OLMo~\citep{groeneveld2024olmo} pipelines can be instrumented simply by wrapping \code{trainer.train()} with \code{with hm.collect()}. The design also naturally accommodates distributed training frameworks such as DDP~\citep{li2020pytorch} and FSDP~\citep{zhao2023pytorch}. Captured gradients may optionally be cached, allowing the same attribution code to score them either online or later from a store persisted during an ordinary training run. Appendix~\ref{app:capture} details the mechanisms underlying these capabilities.

\subsection{Extensibility: pluggable interfaces over a single capture path}
\label{sec:extend}

\dattri{} provides two extension points. The \code{Attributor} abstraction supports the implementation of new gradient-based attribution methods. \code{GradientStreamer} exposes a uniform, well-defined interface to collect per-example gradients. The gradient tier provides the operations required by attribution methods, including inner products, norms, projections, and Kronecker-factored statistics, with each operation routed as described in \S\ref{sec:routing}. A new method can therefore be implemented as a single subclass that consumes these modules and produces the desired attribution scores. The built-in TracIn~\citep{pruthi2020estimating}, K-FAC~\citep{martens2015optimizing}, EK-FAC~\citep{george2018fast,grosse2023studying}, and data-value-embedding~\citep{wang2025capturing} attributors are implemented through this same interface.

The second extension point, \code{HookManagerCallback}, supports downstream applications that operate on gradients during training. A callback receives the capture events emitted at each step and may inspect or modify the training state. Built-in examples include an offloading callback that persists gradients to disk, and a data-selection callback that scores the current batch against a target gradient, and then removes the contributions of low-scoring examples from the parameter gradients before the optimizer step. As demonstrated in \S\ref{sec:dataselect}, this interface reduces the integration of a recent online data-selection method to only a few lines of user code.

\section{Attributing LLMs at scale with high efficiency}
\label{sec:exp-efficiency}

In this section, we empirically evaluate the efficiency of \dattri{} through throughput and memory scaling, and demonstrate the favorable fidelity--cost trade-offs provided by gradient-based attribution methods as model size increases. Supplementary comparisons at a fixed model size and batch size appear in Appendix~\ref{app:fixedbatch}.

\subsection{Scaling attribution with model size}
\label{sec:scaling}

\paragraph{Setup.}
We compare \dattri{}, LogIX~\citep{choe2026your}, Bergson~\citep{quirke2026bergson}, and Kronfluence~\citep{grosse2023studying} on \textsc{GradDot}, K-FAC, and EK-FAC, using Qwen models from $0.5$B to $110$B parameters and $512$-token WikiText-103 sequences~\citep{merity2016pointer} in \code{bf16}. We use factor projection with $k_a=k_g=64$ where supported. Throughput is measured on four H200 GPUs ($141$\,GB each), using each library's supported distributed mode and the largest per-device batch that completes within the benchmark limits. Each run attributes $512$ training sequences, and we use the largest power-of-two batch size that fits in GPU memory. Detailed configurations and timing protocols are reported in Appendix~\ref{app:throughput}.

\begin{figure}[t]
  \centering
  \includegraphics[width=\linewidth]{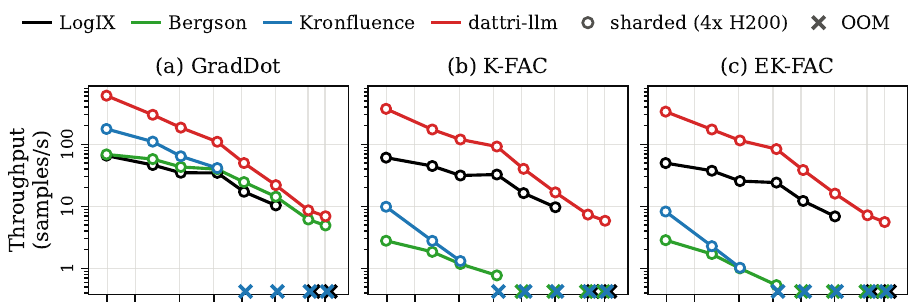}
  \par\vspace{0.3em}
  \includegraphics[width=\linewidth]{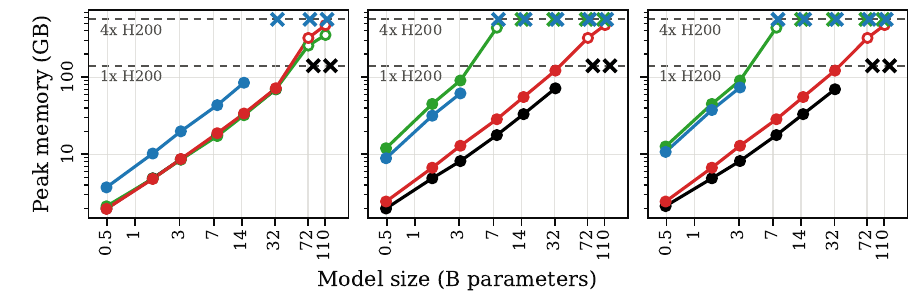}
  \caption{\textbf{Throughput and memory scaling on the Qwen family.} \textbf{Top:} throughput on four H200s at each configuration's largest batch that completes within the benchmark limits. Crosses indicate configurations reported as out of memory. \textbf{Bottom:} GPU memory at batch size one. Filled markers denote single-H200 runs; hollow markers denote four-H200 sharded runs, with reported aggregate memory estimates. Dashed lines indicate aggregate device capacities, and crosses mark the total capacity at which an out-of-memory failure occurs.}
  \label{fig:throughput}
  \vspace{-10pt}
\end{figure}

\paragraph{Results.}
Figure~\ref{fig:throughput} (top) shows that \dattri{} consistently improves attribution throughput across model scales. For \textsc{GradDot}, it achieves $2.7$--$3.5\times$ the throughput of Kronfluence and $1.4$--$8.9\times$ that of Bergson across their respective completed scales. The advantage also persists when compression and batch size are matched: for K-FAC and EK-FAC, \dattri{} achieves on average $3.8\times$ the throughput of LogIX from $0.5$B to $32$B, and up to $6.8\times$ at $0.5$B. The larger gains over Bergson and Kronfluence on curvature-based methods additionally reflect the cost of their full-dimensional curvature factors. Beyond improving throughput, \dattri{} extends the range of feasible workloads: it supports all three methods through $110$B, whereas the evaluated alternatives either exceed the resource budget or lack the sharded execution needed at larger scales.

To understand the memory requirements underlying these scaling results, we also evaluate attribution at batch size one, minimizing batch-dependent memory costs and highlighting the footprint of the model and attribution computation. Figure~\ref{fig:throughput} (bottom) illustrates how the gradient-processing strategy affects scalability. The benefit is most pronounced for curvature-based attribution: at $3$B, \dattri{} uses approximately $17.5$--$20.9\%$ of Kronfluence's GPU memory and $14.2\%$ of Bergson's for K-FAC and EK-FAC. Reducing the attribution-specific memory footprint leaves more capacity for the model and enables larger workloads within the available device memory. Together with sharded execution, this allows \dattri{} to retain support for curvature-based methods at scales where the full-dimensional baseline configurations become infeasible.

\subsection{Attribution fidelity and computational cost}
\label{sec:optfid}

The gradient-based methods supported by \dattri{} approximate the effect of training-data perturbations through first-order sensitivity. We therefore complement the efficiency evaluation with counterfactual retraining, examining whether these approximations provide useful attribution at a practical computational cost.

\paragraph{Setup.}
We evaluate the fidelity and computational cost of attribution methods on training trajectories of GPT-2 ($124$M), Qwen2.5-$1.5$B, and OLMo-3-$7$B~\citep{groeneveld2024olmo}. For each model, we construct $512$ training sequences from WikiText-2 using the model's tokenizer, with $128$ tokens per sequence, and continually pretrain all parameters for one epoch using AdamW with batch size $32$. OLMo-3-$7$B is trained through the unmodified OLMo-core trainer, with \dattri{} capturing gradients by wrapping its \texttt{fit()} call. We assess fidelity against trajectory-specific leave-one-out retraining (TSLOO)~\citep{wang2025capturing}. For each of $50$ randomly selected training sequences, we replay training with that sequence removed and measure the resulting loss changes on $64$ held-out validation sequences. For each validation query, we compute the Spearman correlation between attribution scores and these loss changes across the $50$ training sequences, then average over queries. Training and evaluation use \code{fp32} with \code{TF32} and dropout disabled to resolve the small counterfactual loss changes.

We compare AdamW-influence~\citep{deng2026faithfultrajectorybaseddataattribution} and EK-FAC implemented in \dattri{} with Bergson's MAGIC~\citep{ilyas2025magic}, SOURCE~\citep{bae2024source}, TrackStar~\citep{chang2024trackstar}, and EK-FAC~\citep{grosse2023studying}. For AdamW-influence, we evaluate full-dimensional gradients and reduced representations that average scores from ten disjoint random masks of $k\in\{512,8192\}$ gradient coordinates per layer. For EK-FAC, we evaluate full-dimensional factors and projected factors with $k_a=k_g=64$. Each execution receives one B200, $256$\,GB of host memory, and $1$\,TB of local disk. Attribution throughput measures the number of queries scored per second, excluding trajectory training, model and data loading, and TSLOO reference construction. Appendix~\ref{app:fidelity} provides complete settings, results, and resource limits.

\begin{figure}[t]
  \centering
  \includegraphics[width=\linewidth]{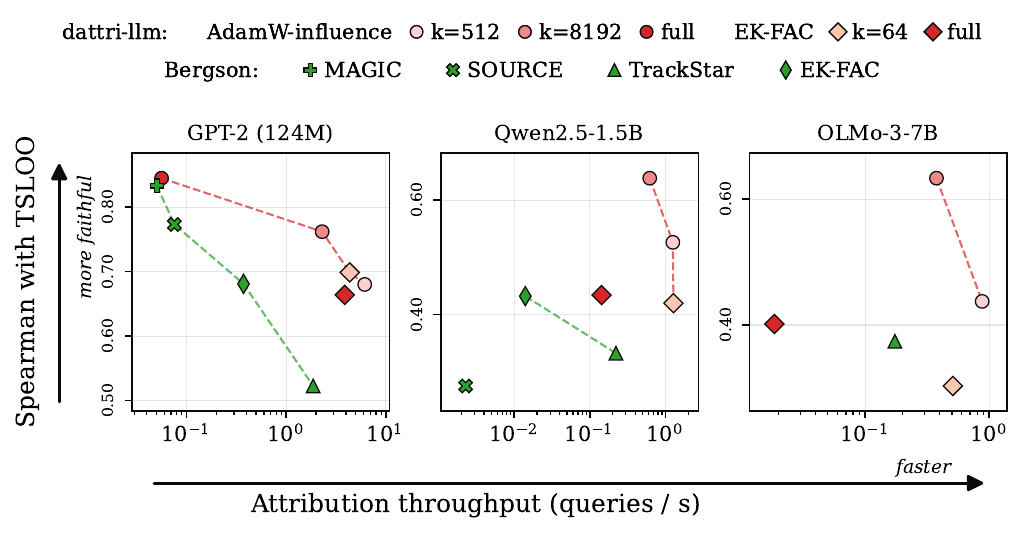}
  \vspace{-20pt}
  \caption{\textbf{Attribution fidelity versus throughput.} Higher and farther right is better. Fidelity is measured as the mean query-wise Spearman correlation with TSLOO over $64$ validation queries. Throughput is $64$ divided by the attribution time in seconds on one B200. Dashed lines show the empirical Pareto frontier for each library. Configurations exceeding the resource budget are omitted.}
  \label{fig:scaling-fidelity}
  \vspace{-10pt}
\end{figure}

\paragraph{Results.}
Figure~\ref{fig:scaling-fidelity} shows that \dattri{} offers a favorable fidelity--cost trade-off across all three models, with its configurations forming the empirical Pareto frontier among completed runs in two of the three panels. On GPT-2, full-dimensional AdamW-influence achieves a correlation of $0.845$, compared with MAGIC's $0.833$, while requiring less time. Masking and projection provide further control over this trade-off: for AdamW-influence, increasing the mask size $k$ from $512$ to $8192$ improves correlation across all three models at the cost of approximately $2$--$3\times$ longer attribution time. Beyond attribution efficiency, \dattri{}'s non-invasive gradient capture enables end-to-end attribution for OLMo-3-$7$B trained with the unmodified OLMo-core trainer, demonstrating compatibility with an existing training pipeline at larger scale.

\section{Integrating and extending attribution workflows}
\label{sec:dataselect}

We further demonstrate how the interfaces of \S\ref{sec:extend} make a downstream application easy to build, highlighting the compatibility and extensibility of \dattri{}. We take the most recent attribution-based data selection technique for post-training~\citep{hu2026dr} and deploy it inside a minimal fine-tuning loop. Specifically, at each training step we compute \textsc{GradDot} scores between the samples in the current batch and a held-out validation set, and drop the lowest-scoring samples from the gradient update.

\paragraph{Setup.}
Following \citet{hu2026dr}, we fine-tune Llama-$3.2$-$1$B~\citep{grattafiori2024llama} on Alpaca~\citep{taori2023alpaca} with SAMSum~\citep{gliwa2019samsum} as the target task, dropping the bottom half of each batch at every step, and compare three variants: \emph{Full-Training}, which trains without data selection; \emph{Global Subset}, which ranks samples by \textsc{GradDot} scores aggregated over all layers; and \emph{Layer-Wise Subset}, which ranks samples independently for each layer. All variants are run under both full-parameter and LoRA~\citep{hu2022lora} fine-tuning. Hyperparameters are reported in Appendix~\ref{app:dataselect}. Figure~\ref{fig:dataselect-code} shows the core integration, which reduces to three steps. First, the selection policy is expressed as a \code{DataSelectionCallback} (\S\ref{sec:extend}), which specifies the target loader and the selection rule. Second, the callback is attached to a \code{HookManager}, which captures per-sample gradients during the forward and backward passes at almost no additional cost. Third, the user's training loop is enclosed in \code{hm.collect()}, with the loop itself unchanged. Between the backward pass and the optimizer step, the callback scores the batch and removes the dropped samples' contributions from the parameter gradients. The same procedure applies to LoRA fine-tuning, which differs from full-parameter fine-tuning only in a layer selector that restricts the hooks to the adapters.

\begin{figure}[t]
  \centering
  \small
  { SFT with Data Selection: \quad \textcolor[HTML]{000000}{\raisebox{0.45ex}{\rule{1.8em}{1pt}}}~Full-Training\quad \textcolor[HTML]{2CA02C}{\raisebox{0.45ex}{\rule{1.8em}{1pt}}}~Global Subset\quad \textcolor[HTML]{D62728}{\raisebox{0.45ex}{\rule{1.8em}{1pt}}}~Layer-Wise Subset }
  \vspace{0.4em}
  \includegraphics[width=\linewidth]{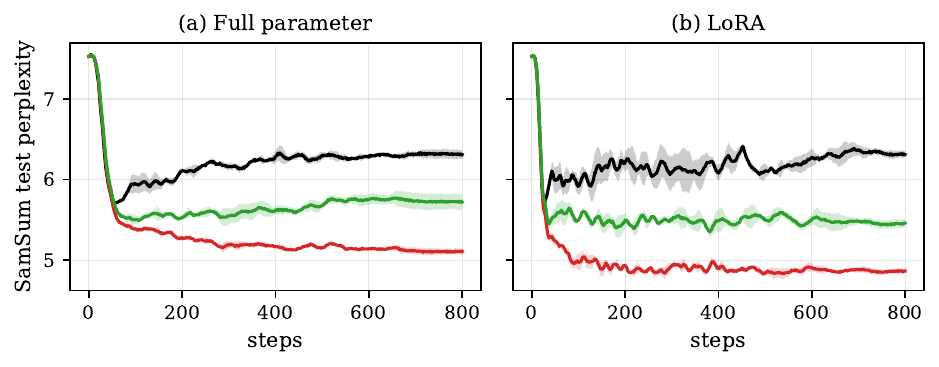}
  \vspace{-25pt}
  \caption{Attribution-guided data selection under \textbf{(a)} full-parameter and \textbf{(b)} LoRA fine-tuning (SAMSum test perplexity, lower is better; mean over $3$ seeds, band $\pm 1$ std). Both selection variants improve on training with all data, with per-layer selection best in both regimes.}
  \label{fig:dataselect}
  \vspace{-10pt}
\end{figure}

\begin{wrapfigure}[14]{r}{0.56\textwidth}
\vspace{-15pt}
\begin{lstlisting}[style=dattripy]
from dattri_llm import DataSelectionCallback, HookManager
# 1. Define the selection policy
select = DataSelectionCallback(
    model, target="val_loader",
    val_loader=target_loader, val_loss_fn=loss_fn,
    selection_kwargs={"threshold": 0.5, "threshold_mode": "bottom_fraction"})
# 2. Define the capture hook
hm = HookManager(model, callbacks=[select])
# 3. Perform data selection
with hm.collect():
    for batch in train_loader:
        loss_fn(model, batch).backward()
        optimizer.step(); optimizer.zero_grad()
\end{lstlisting}
\caption{Online data selection with \dattri{}.}
\label{fig:dataselect-code}
\end{wrapfigure}

\paragraph{Results.}
Figure~\ref{fig:dataselect} shows the test perplexity curves. Both selection variants outperform Full-Training in both regimes, and Layer-Wise Subset performs best, reaching a final target perplexity of $5.11$ vs.\ $6.31$ under full-parameter fine-tuning and $4.87$ vs.\ $6.31$ under LoRA. Full-Training overfits the off-target source data and its perplexity drifts upward, whereas selection acts as a regularizer that steers each update toward target-aligned examples, and the finer-grained per-layer selection regularizes further. The result reproduces the finding of \citet{hu2026dr} using the integration of Figure~\ref{fig:dataselect-code} in place of a dedicated training pipeline. Notably, the \dattri{} implementation incurs negligible runtime overhead compared with the original implementation by \citet{hu2026dr}. We report runtime measurements in Appendix~\ref{app:dataselect}.

\section{Conclusion}
\label{sec:conclusion}

We presented \dattri{}, a library for training data attribution at LLM scale, designed for efficiency, compatibility, and extensibility. For efficiency, it represents per-example gradients exactly in factorized or materialized form and uses a FLOP-aware cost model to select the cheaper route, sharing kernels across attribution methods and supported layers without changing scores. For compatibility, non-invasive autograd hooks capture gradients from unmodified training pipelines, including DDP and FSDP. For extensibility, a callback interface supports applications such as online data selection without modifying the core. Empirically, \dattri{} is $3.2\times$ faster than the fastest existing libraries on average. We believe \dattri{} can lower the systems barrier to studying how training data shapes large language models.

\subsection*{AI use statement}

In this work, we used generative AI tools for implementing methods. We have not used generative AI tools to help develop theoretical models or conceptual frameworks, formulate mathematical claims, provide critical ingredients for proving mathematical claims, assist in the writing of proofs, propose or refine hypotheses, design or provide feedback on research methodology or experiments, assist with translation, clean and reformat dataset, or interpret results, and Generate synthetic data sets and support qualitative and thematic data analysis are not applicable to this work. Additionally, we used generative AI tools to create or modify scientific figures or images, draft parts of a research paper, edit a research paper to improve readability, and format references. We have reviewed all AI-assisted work. The authors reviewed AI-assisted implementations, figures, text, and references for correctness and consistency with the methods and experimental results. We take responsibility for the final content of this work, including text, claims or artifacts produced with the aid of generative AI.

\subsection*{Reproducibility statement}

We document the library design and experimental procedures to support reproducibility. All experimental details are included in Appendix~\ref{app:experiment}. Appendix~\ref{app:throughput} specifies the detailed setup for the throughput measurement experiment, including the hardware, model checkpoints, workloads, distributed configurations, batch-size selection, and timing and memory measurement protocols. Appendix~\ref{app:fidelity} details the fidelity evaluation, including training settings, counterfactual retraining, numerical controls, and evaluation metrics. Appendices~\ref{app:fixedbatch}, \ref{app:routingexp}, and \ref{app:tokenmore} provide the settings for matched-batch comparisons, routing ablations, and token-level attribution examples. Appendix~\ref{app:dataselect} documents the datasets, hyperparameters, selection procedure, and evaluation protocol for online data selection, with results reported over three random seeds. We will officially release the code upon acceptance of the paper.

\bibliography{references}
\bibliographystyle{iclr2027_conference}

\appendix
\section{Design Details}
\label{app:design}

This appendix expands the design of \S\ref{sec:design}: the canonical factorization for non-linear layer types (\S\ref{app:layers}), the dimension-reduction maps and how they are applied to the factors (\S\ref{app:projection}), optimizer-preconditioned capture (\S\ref{app:optcapture}), the routing cost model (\S\ref{app:routing}), the capture mechanism (\S\ref{app:capture}), and the storage and streaming interfaces that connect capture to attribution (\S\ref{app:storage}).

\subsection{A canonical factorization for all supported layers}
\label{app:layers}

Equation~\ref{eq:canonical-factorization} is not specific to linear layers. For every supported layer type, \dattri{} provides an \emph{unfolding map} that reshapes the layer's parameters into a matrix $W\in\mathbb{R}^{N_o\times N_i}$ and expresses its flattened gradient $\gamma=\operatorname{vec}(\mathrm{D}W)$ in the same form, with $a_t$ an unfolded input vector, $g_t$ the corresponding output gradient, and the sum running over $T'$ unfolded positions in place of $T$. Convolutional layers unfold input patches (\texttt{im2col}); transposed convolutions use the same unfolding with the roles of input and output channels exchanged; embedding layers treat the one-hot token id as the input factor, so that materialization scatters $g_t$ into the row of the token; and normalization layers (\code{LayerNorm}, \code{RMSNorm}, \code{GroupNorm}, \code{InstanceNorm}) use the normalized activation as the input factor, with the bias folded into an extra ones column where present. The supported types are \code{nn.Linear} (including HuggingFace \code{Conv1D}), \code{nn.Conv1d/2d/3d}, \code{nn.ConvTranspose1d/2d/3d}, \code{nn.Embedding} and \code{nn.EmbeddingBag}, and the normalization modules above. Because every operation in the library depends on a layer only through Eq.~\ref{eq:canonical-factorization}, a single implementation of each operation, and hence of each attribution method, covers all of them. Trainable parameters outside these families can still be captured through a batch-level parameter-gradient hook, which yields the materialized form directly.

\subsection{Dimension reduction}
\label{app:projection}

\S\ref{sec:postproc} introduces factor projection (Eq.~\ref{eq:factor-projection}) and materialized projection. This section records how each is applied and which random maps are available.

\paragraph{Factor projection.}
The projection matrices $P_a$ and $P_g$ are applied inside the backward hook, so the raw factors are never buffered and only the projected pair $(\tilde a_t,\tilde g_t)$ is retained. Because the result has the form of Eq.~\ref{eq:canonical-factorization} with $(N_i,N_o)$ replaced by $(k_a,k_g)$, every operation defined on factorized gradients, including routing (\S\ref{app:routing}) and the Kronecker-factored statistics, applies to the projected factors without change. When $k_a=k_g$ the construction coincides with the convention of LoGRA~\citep{choe2026your} and GraSS~\citep{hu2026grass}. The two sides use distinct seeds derived from a per-layer base seed, which must be held fixed across every gradient that will be compared, including gradients captured at different steps or on different ranks.

\paragraph{Materialized projection.}
A single matrix $P\in\mathbb{R}^{k_{\text{dense}}\times(N_iN_o)}$ is applied to the flattened gradient, $\tilde\gamma=P\gamma\in\mathbb{R}^{k_{\text{dense}}}$, as in TRAK~\citep{park2023trak}. For normalization layers, materialized projection is the only projection available. Their factors are captured like those of any other layer, but their weight gradient is an elementwise product of the two factors summed over positions, $\sum_{t}\hat{x}_t\odot g_t$, rather than an outer product, so factor projection does not apply; the dense gradient, with one entry per feature, is also smaller than the factors it is computed from. Materialized projection is incompatible with the Kronecker-factored methods, which require the factors.

\paragraph{Random maps.}
Either projection may use a dense Gaussian map~\citep{johnson1984extensions}, a Rademacher map with entries $\pm 1/\sqrt{d_{\mathrm{out}}}$, where $d_{\mathrm{out}}$ is the map's output dimension ($k_a$, $k_g$, or $k_{\text{dense}}$)~\citep{achlioptas2003database}, a sparse Johnson--Lindenstrauss transform with a fixed number of nonzeros per column~\citep{kane2014sparser}, or the sparsify-then-project map of GraSS~\citep{hu2026grass}. In addition, a \emph{random mask} keeps a fixed random subset $\mathcal{M}$ of the entries of $\mathrm{D}W$. Because entry $(r,c)$ of $\mathrm{D}W$ equals $\sum_{t=1}^{T}g_{t,r}\,a_{t,c}$, the masked entries are gathered directly from the factors as a sum of $|\mathcal{M}|$ products per token position, without forming $\mathrm{D}W$. The mask is the one reduction that preserves the coordinate identity of each retained entry, which is what optimizer preconditioning (\S\ref{app:optcapture}) requires. All maps are generated from the same per-layer seed so that projected gradients from different steps, ranks, and passes remain comparable.

\subsection{Optimizer-preconditioned capture}
\label{app:optcapture}

Modern models are trained with AdamW rather than SGD, and a growing family of attribution methods scores the \emph{update} a training example induces rather than its raw gradient: LESS~\citep{xia2024less} compares the Adam direction of an example with a query gradient, and AdamW-influence~\citep{deng2026faithfultrajectorybaseddataattribution} unrolls that direction through the remaining AdamW steps. For Adam-family optimizers, Eq.~\ref{eq:adam-direction} maps sample $i$'s flattened gradient $\gamma^{(i)}$ to its update direction $\Gamma^{(i)}$ at optimizer step $\tau$, using the moment estimates $(m,v)$ before the coming \code{optimizer.step()}. Eq.~\ref{eq:adam-direction} has properties that shape the design. It needs the optimizer state \emph{at the step the example was used}, so it must be evaluated during training; and it is nonlinear in $\gamma^{(i)}$ coordinate by coordinate, so it does not commute with a random projection, which mixes coordinates.

\dattri{} applies optimizer preconditioning to gradient entries recovered from the factors. \code{HookManager} accepts the training optimizer, and for each hooked layer the per-example gradient entries are recovered from the factors and mapped through the optimizer's update rule before they are buffered, so only the preconditioned copy ever exists, at the same place and with the same cost profile as factor projection. The random mask of \S\ref{app:projection} pairs naturally with it: the mask selects the coordinates, the entries are recovered from the factors, and the rule is applied to exactly those entries. Every coordinate-wise optimizer in \code{torch.optim} is supported through one kernel, and the map is switchable per pass so that a raw query pass can share the hooks of a preconditioned training pass. For trajectory methods a callback additionally records the moments on the same coordinates before and after every update, from which the batch gradient the optimizer consumed is recovered exactly. On this capture path, LESS is TracIn over preconditioned gradients with a cosine metric, in either its per-step or its checkpoint form, and AdamW-influence is a backward recurrence over the stored steps. The recurrence is linear in its summary matrix, so it can be carried either on the training side, at $O(|\mathcal{M}|^2)$ memory in the mask size $|\mathcal{M}|$, or on the query side through the query gradients, which removes the mask entirely at a cost linear in the number of queries (\S\ref{sec:optfid}).

\subsection{FLOP-aware routing}
\label{app:routing}

Consider the cross-Gram matrix of gradient inner products between two batches of sizes $B_1$ and $B_2$, with $T$ tokens per example and layer dimensions $N_i$, $N_o$. Table~\ref{tab:routing-complexity} summarizes the factorized and materialized routes.

\begin{table}[h]
    \centering
    \small
    \caption{Complexity of computing a cross-Gram matrix between two batches of per-example gradients. Space denotes the storage required for the gradient representations and excludes the $O(B_1B_2)$ output matrix common to both routes. Under factor projection (Eq.~\ref{eq:factor-projection}), $N_i$ and $N_o$ are replaced by the projection dimensions $k_a$ and $k_g$.}
    \label{tab:routing-complexity}
    \begin{tabular}{lll}
        \toprule
        Route & Arithmetic operations & Representation space \\
        \midrule
        Factorized & $O\!\left(B_1B_2T^2(N_i+N_o)\right)$
                   & $O\!\left((B_1+B_2)T(N_i+N_o)\right)$ \\
        Materialized & $O\!\left((B_1+B_2)TN_iN_o+B_1B_2N_iN_o\right)$
                     & $O\!\left((B_1+B_2)N_iN_o\right)$ \\
        \bottomrule
    \end{tabular}
\end{table}

\paragraph{Regimes.}
The materialized route pays a one-time construction cost of $TN_iN_o$ operations per example, amortized across all $B_1B_2$ pairwise comparisons of cost $N_iN_o$ each; the factorized route avoids construction but pays $T^2(N_i+N_o)$ per pair. Short sequences and wide layers therefore favor the factorized route, and long sequences or large comparison batches favor materialization. The storage comparison has a closed form: the factorized form is smaller exactly when $T<N_iN_o/(N_i+N_o)$. For unprojected wide layers this threshold is on the order of the layer width, so factors are kept; under factor projection it shrinks to $k_ak_g/(k_a+k_g)$, which realistic sequence lengths far exceed, so projected factors are better materialized as small $k_g\times k_a$ gradient matrices. The projected and unprojected regimes thus select opposite representations under the same rule.

\paragraph{Routing.}
Routing is evaluated per layer and per operation, for cross-Gram matrices and for per-example norms alike, by comparing the operation counts above and taking the cheaper route; the same routed kernels serve every attribution method and the data-selection callback. At capture time the representation that is persisted follows the storage rule and is exposed as a per-layer setting. Because the routes are algebraically equivalent, routing does not change attribution scores in exact arithmetic. Appendix~\ref{app:routingexp} analyzes the case of reused query representations and evaluates routing across sequence lengths.

\subsection{Capture mechanism}
\label{app:capture}

\paragraph{Hook families.}
\code{HookManager} attaches a hook family to each selected layer. The default \emph{factorized} family registers a forward hook that records the layer input and a full backward hook that records the output gradient. A \emph{parameter-gradient} family registers a tensor hook on each parameter and records the batch-level gradient in materialized form; it is also the fallback for trainable layers outside the supported types. Layers are selected by type, by regular expression, or by an all-eligible marker, and layers whose type is not recognized automatically, such as custom normalization modules, are declared through per-layer overrides rather than by editing the model. When the training call is buried inside a framework, the capture context can be opened and closed from a framework callback instead of a \code{with} block.

\paragraph{Loop-independent step detection.}
Because the manager never sees the training loop, it infers when a step has finished from the hooks themselves. It tracks which hooked layers participated in the current forward pass and waits until every one has fired its backward hook, with an additional end-of-backward barrier for parameter gradients; layers skipped by the current control path are not waited for. Repeated invocations of a layer within a step (weight tying, multiple forward calls) are tracked as separate virtual layers. Under gradient checkpointing, the recomputed forward pass is recognized by the presence of a live autograd task and matched to the correct backward pass rather than double-counted. For backward passes that must not contribute, such as a validation pass used to form a target gradient, the manager can save, clear, and restore its per-step state around the extra pass, so the training step is captured exactly as if the extra pass had not occurred.

\paragraph{Parallelism.}
Under DDP~\citep{li2020pytorch}, parameter hooks fire when a parameter's local gradient is computed, before the all-reduce, so each rank captures the gradient of its own batch; under FSDP~\citep{zhao2023pytorch}, per-parameter gradient hooks do not fire, so step completion is signaled through an end-of-backward callback queued on the autograd engine. In both cases each rank hooks its local shard and writes records independently, and the storage manager merges the per-rank indexes at read time, so sharded capture reproduces single-device gradients. The data-selection callback's edit of \code{param.grad} is collective: contributions are all-reduced and applied to the full tensor under DDP or to the owned shard under FSDP, and under DDP the edit is queued to run after the reducer's own write-back.

\paragraph{Integration with training frameworks.}
Figure~\ref{fig:trainer-integration} shows the capture code for the HuggingFace \code{Trainer}, TRL's \code{GRPOTrainer}, and the OLMo trainer. In each case the trainer is constructed and called exactly as without attribution: the user builds a \code{HookManager} over the model, attaches callbacks that offload the captured gradients, and runs the unmodified training call inside \code{hm.collect()}. The optional layer selectors are provided and can be tailored for different model architectures.

\begin{figure}[t]
\centering
\small
\textbf{(a) HuggingFace \code{Trainer}}
\begin{lstlisting}[style=dattripy, moreemph={OffloadCallback,GradientStorageManager}]
from dattri_llm import (GradientStorageManager, HookManager,
                        HookManagerConfig, OffloadCallback)
store = GradientStorageManager("gradients/")
hm = HookManager(model,
    config=HookManagerConfig(linear_io=[r"transformer\.h\.", r"wte", r"lm_head"]),
    callbacks=[OffloadCallback(offload_interval=1, file_manager=store)])
# Pattern A: wrap the unmodified training call
trainer = Trainer(model=model, args=training_args, train_dataset=dataset)
with hm.collect():
    trainer.train()
# Pattern B: open and close capture from a Trainer callback
class CollectCallback(TrainerCallback):
    def __init__(self, hm):
        self.hm, self.stack = hm, contextlib.ExitStack()
    def on_train_begin(self, args, state, control, **kwargs):
        self.stack.enter_context(self.hm.collect())
    def on_train_end(self, args, state, control, **kwargs):
        self.stack.close()
trainer = Trainer(model=model, args=training_args, train_dataset=dataset,
                  callbacks=[CollectCallback(hm)])
trainer.train()
\end{lstlisting}
\vspace{2pt}
\textbf{(b) TRL \code{GRPOTrainer}}
\begin{lstlisting}[style=dattripy, moreemph={OffloadCallback,GradientStorageManager}]
trainer = GRPOTrainer(model=model, reward_funcs=reward_fn, args=grpo_config,
                      train_dataset=prompts, processing_class=tokenizer)
hm = HookManager(model,
    config=HookManagerConfig(linear_io=[r"transformer\.h\."]),
    callbacks=[OffloadCallback(offload_interval=1, file_manager=store)])
with hm.collect():
    trainer.train()
\end{lstlisting}
\vspace{2pt}
\textbf{(c) OLMo \code{Trainer}}
\begin{lstlisting}[style=dattripy, moreemph={OffloadCallback,GradientStorageManager}]
trainer = Trainer(cfg=cfg, model=model, dist_model=dist_model, optim=optim,
                  scheduler=scheduler, train_loader=loader, device=device,
                  evaluators=[])
hm = HookManager(model,
    config=HookManagerConfig(linear_io=[r"ff_proj", r"ff_out"]),
    callbacks=[OffloadCallback(offload_interval=1, file_manager=store)])
with hm.collect():
    trainer.fit()
\end{lstlisting}
\caption{Gradient capture around three training frameworks. The trainer and its training call are unmodified.}
\label{fig:trainer-integration}
\end{figure}

\subsection{Storage, identity, and the capture--attribution interface}
\label{app:storage}

\paragraph{Sample identity.}
Offloaded gradients are keyed by a content hash of a sample's model inputs, which is independent of position and shuffling. The on-disk index maps \code{input\_hash} $\to$ \code{(step, sample\_idx)} $\to$ file, so retrieving one sample's gradient is a direct slice rather than a scan of the batch record it was written in.

\paragraph{Store.}
Records are written before they are indexed and the index is persisted atomically, so an interrupted run leaves the store readable up to the last indexed record rather than corrupt. Under distributed training, each rank writes to its own subdirectory and a freshly constructed manager merges the per-rank indexes, so the number of ranks used for capture is invisible to the reader. The offload callback can merge the micro-batches of a gradient-accumulation window into one record stamped with the optimizer step, which is exact because parameters do not change within a window.

\paragraph{Streaming interface.}
Attributors consume gradients through one contract: a source yielding per-step \code{(step, gradient, hashes)} blocks. A disk-backed source re-iterates over pre-collected gradients, loading one file per item so a standard data loader can prefetch whole blocks. A live source computes blocks from a forward and backward pass, either as a re-iterable probe at a fixed checkpoint or, in its updating mode, advancing the optimizer so each step is a genuine point on a training trajectory; the updating mode mirrors the HuggingFace \code{Trainer}'s inner loop (optimizer and scheduler construction, mixed precision, clipping, and distributed wrapping). Attributors branch on exactly one property of the source, whether it is re-iterable, which methods with a pre-pass such as K-FAC's Fisher fit require. Scores are returned as an \code{AttributionScore} whose rows are (example, step) pairs, so trajectory-aware and trajectory-agnostic methods share one container.

\paragraph{Workflows.}
The same attributor code serves the \emph{on-the-fly} workflow, where an attributor drives a live source over the model and data and scores in one run, and the \emph{store-then-attribute} workflow, where an ordinary training run caches gradients via the offload callback and attribution is performed later over the store, without the model and without another backward pass, so different methods and settings can be re-run over the same cache.

\section{Experiment Details}
\label{app:experiment}

\subsection{Throughput Under a Fixed Hardware Budget}
\label{app:throughput}

We evaluate the Qwen family of models in Table~\ref{tab:scaling-models} on WikiText-103 with sequence length $512$ and bf16 precision. Each configuration scores $512$ training sequences against one query through $32$B and four queries at $72$B and $110$B. The query count is matched across libraries at each scale. Each run receives four H200 GPUs ($141$\,GB per device), $32$ CPU cores, and $256$\,GB of host memory. Attention and MLP linear layers are included in attribution; embeddings and the language-model head are excluded.

\begin{table}[t]
  \centering
  \small
  \setlength{\tabcolsep}{4pt}
  \caption{Qwen models used in the scaling experiments: parameters (B), transformer blocks $L$, model width $d$, MLP width $d_{\text{ff}}$, attention heads (query / key-value), and hooked linear layers ($7L$). The Qwen family provides a smooth progression in model size from $0.5$B to $110$B.}
  \label{tab:scaling-models}
  \begin{tabular}{llrrrrrr}
    \toprule
    Scale & Checkpoint & Params & $L$ & $d$ & $d_{\text{ff}}$ & Heads & Hooked \\
    \midrule
    $0.5$B & Qwen2.5-0.5B & $0.49$ & $24$ & $896$  & $4864$  & $14/2$ & $168$ \\
    $1.5$B & Qwen2.5-1.5B & $1.54$ & $28$ & $1536$ & $8960$  & $12/2$ & $196$ \\
    $3$B   & Qwen2.5-3B   & $3.09$ & $36$ & $2048$ & $11008$ & $16/2$ & $252$ \\
    $7$B   & Qwen2.5-7B   & $7.62$ & $28$ & $3584$ & $18944$ & $28/4$ & $196$ \\
    $14$B  & Qwen2.5-14B  & $14.8$ & $48$ & $5120$ & $13824$ & $40/8$ & $336$ \\
    $32$B  & Qwen2.5-32B  & $32.5$ & $64$ & $5120$ & $27648$ & $40/8$ & $448$ \\
    $72$B  & Qwen2.5-72B  & $72.7$ & $80$ & $8192$ & $29568$ & $64/8$ & $560$ \\
    $110$B & Qwen1.5-110B & $111$  & $80$ & $8192$ & $49152$ & $64/8$ & $560$ \\
    \bottomrule
  \end{tabular}
\end{table}

\paragraph{Library configurations.}
\dattri{}, Bergson and Kronfluence use FSDP. LogIX uses one model replica per device under DDP. Factor projection with $k_a=k_g=64$ is requested where supported. Kronfluence runs without that projection. Consequently, the throughput comparison reflects the capabilities of the evaluated implementations as well as their execution efficiency. For each library, method, and model size, we manually test power-of-two per-device batch sizes and select the largest that can successfully complete the workload.

\paragraph{Timing.}
After two untimed warm-up steps, the GPU-synchronized timed interval covers the complete attribution computation: gradient capture, Kronecker-factor estimation and eigendecomposition where required, and scoring. Model loading, data preparation, and process initialization are excluded. For Bergson, which launches workers for successive pipeline stages, the reported time further removes the measured worker initialization and model-loading intervals for fairness. Throughput is $512$ divided by the resulting elapsed time; it measures the complete finite workload and includes necessary per-call overheads for each attribution method.

\paragraph{Memory measurements.}
Table~\ref{tab:throughput-memory} reports peak per-device memory at the selected batch size. For \dattri{}, LogIX, and Kronfluence we use \code{torch.cuda.max\_memory\_allocated}; Bergson uses NVML readings because its computation runs in child processes. The memory comparison at a common batch size is further reported below.

\begin{table}[t]
  \centering\footnotesize
  \setlength{\tabcolsep}{3pt}
  \caption{Throughput (training sequences/s) on four H200s, with the per-device batch size in parentheses. The workload contains $512$ training sequences and one query through $32$B, and four queries at $72$B and $110$B. A dash denotes a configuration reported as out of memory.}
  \label{tab:throughput-full}
  \begin{tabular}{llrrrrrrrr}
    \toprule
    Method & Library & $0.5$B & $1.5$B & $3$B & $7$B & $14$B & $32$B & $72$B & $110$B \\
    \midrule
    \multirow{4}{*}{\textsc{GradDot}} & \dattri{} & $613$ (128) & $299$ (64) & $186$ (32) & $110$ (32) & $49.9$ (16) & $22.0$ (8) & $8.6$ (8) & $6.9$ (4) \\
     & Bergson & $68.8$ (64) & $57.9$ (64) & $43.2$ (32) & $39.8$ (32) & $24.6$ (16) & $14.3$ (8) & $6.1$ (8) & $4.9$ (4) \\
     & Kronfluence & $177$ (64) & $111$ (32) & $64.4$ (16) & $41.5$ (16) & --- & --- & --- & --- \\
     & LogIX & $65.8$ (128) & $46.5$ (64) & $34.8$ (32) & $34.7$ (32) & $17.2$ (16) & $10.4$ (8) & --- & --- \\
    \midrule
    \multirow{4}{*}{K-FAC} & \dattri{} & $375$ (128) & $174$ (64) & $120$ (32) & $92.2$ (32) & $40.3$ (16) & $16.8$ (8) & $7.4$ (8) & $5.9$ (4) \\
     & Bergson & $2.8$ (32) & $1.9$ (32) & $1.2$ (16) & $0.8$ (8) & --- & --- & --- & --- \\
     & Kronfluence & $9.9$ (64) & $2.8$ (32) & $1.3$ (16) & --- & --- & --- & --- & --- \\
     & LogIX & $61$ (128) & $44.8$ (64) & $31.5$ (32) & $32.6$ (32) & $16.4$ (16) & $9.7$ (8) & --- & --- \\
    \midrule
    \multirow{4}{*}{EK-FAC} & \dattri{} & $337$ (128) & $174$ (64) & $116$ (32) & $83.9$ (32) & $38.6$ (16) & $16.1$ (8) & $7.2$ (8) & $5.6$ (4) \\
     & Bergson & $2.9$ (32) & $1.7$ (16) & $1$ (8) & $0.5$ (4) & --- & --- & --- & --- \\
     & Kronfluence & $8.3$ (64) & $2.3$ (32) & $1$ (8) & --- & --- & --- & --- & --- \\
     & LogIX & $49.9$ (128) & $37.5$ (64) & $25.6$ (32) & $24.2$ (32) & $12.2$ (16) & $6.9$ (8) & --- & --- \\
    \bottomrule
  \end{tabular}
\end{table}

\begin{table}[t]
  \centering
  \setlength{\tabcolsep}{3pt}
  \caption{Peak per-device GPU memory (GB) at the batch sizes used in Table~\ref{tab:throughput-full}. Batch sizes vary across methods, libraries, and scales; these values document the throughput configurations and do not constitute a fixed-batch memory comparison.}
  \label{tab:throughput-memory}
  \begin{tabular}{llrrrrrrrr}
    \toprule
    Method & Library & $0.5$B & $1.5$B & $3$B & $7$B & $14$B & $32$B & $72$B & $110$B \\
    \midrule
    \multirow{4}{*}{\textsc{GradDot}} & \dattri{} & $117.6$ & $93.0$ & $67.2$ & $88.4$ & $72.0$ & $79.0$ & $131.0$ & $127.8$ \\
     & Bergson & $80.9$ & $119.0$ & $86.5$ & $104.6$ & $87.7$ & $91.2$ & $139.8$ & $135.0$ \\
     & Kronfluence & $101.7$ & $95.3$ & $80.7$ & $119.7$ & --- & --- & --- & --- \\
     & LogIX & $117.4$ & $94.6$ & $70.9$ & $97.4$ & $91.0$ & $124.4$ & --- & --- \\
    \midrule
    \multirow{4}{*}{K-FAC} & \dattri{} & $117.6$ & $93.0$ & $67.2$ & $88.4$ & $72.0$ & $80.0$ & $132.1$ & $127.8$ \\
     & Bergson & $95.4$ & $134.4$ & $112.4$ & $132.5$ & --- & --- & --- & --- \\
     & Kronfluence & $101.7$ & $95.3$ & $98.6$ & --- & --- & --- & --- & --- \\
     & LogIX & $117.4$ & $94.6$ & $70.9$ & $96.8$ & $91.0$ & $124.4$ & --- & --- \\
    \midrule
    \multirow{4}{*}{EK-FAC} & \dattri{} & $117.6$ & $93.0$ & $67.2$ & $88.4$ & $72.0$ & $80.0$ & $132.1$ & $127.8$ \\
     & Bergson & $132.5$ & $121.2$ & $109.2$ & $134.1$ & --- & --- & --- & --- \\
     & Kronfluence & $108.7$ & $122.6$ & $100.4$ & --- & --- & --- & --- & --- \\
     & LogIX & $117.4$ & $94.3$ & $71.4$ & $97.4$ & $91.0$ & $124.4$ & --- & --- \\
    \bottomrule
  \end{tabular}
\end{table}

\paragraph{Memory Scaling at Batch Size One}
We evaluate GPU memory usage at batch size one to assess the largest model each library can accommodate, using the same models and workload as above. Each configuration is evaluated on one H200; when a single device is insufficient, we use four H200s if the library supports sharding. LogIX does not support sharding in its current implementation. Table~\ref{tab:scaling-full} explicitly marks completed four-device sharded runs and reports their estimated aggregate memory usage across all four devices. A dash indicates that the configuration runs out of memory under this evaluation protocol.

\begin{table}[t]
  \centering\small
  \setlength{\tabcolsep}{5pt}
  \caption{GPU memory usage (GB) at batch size one, with $64$ training sequences and one query. $^{\mathrm{F}}$ denotes a completed sharded run on four H200s, with memory reported as an estimated aggregate across all four devices. All other completed runs use one H200. A dash indicates an out-of-memory failure under the evaluation protocol.}
  \label{tab:scaling-full}
  \begin{tabular}{llrrrr}
    \toprule
    Method & Scale & \dattri{} & Bergson & LogIX & Kronfluence \\
    \midrule
    \multirow{8}{*}{\textsc{GradDot}} & $0.5$B & $2$ & $2.1$ & $2$ & $3.7$ \\
     & $1.5$B & $4.8$ & $4.8$ & $4.9$ & $10.2$ \\
     & $3$B & $8.7$ & $8.5$ & $8.6$ & $19.9$ \\
     & $7$B & $18.8$ & $17.3$ & $17.8$ & $43.5$ \\
     & $14$B & $33.8$ & $32.2$ & $33.2$ & $85$ \\
     & $32$B & $72.1$ & $69.9$ & $69.8$ & --- \\
     & $72$B & $323^{\mathrm{F}}$ & $258^{\mathrm{F}}$ & --- & --- \\
     & $110$B & $476^{\mathrm{F}}$ & $354^{\mathrm{F}}$ & --- & --- \\
    \midrule
    \multirow{8}{*}{K-FAC} & $0.5$B & $2.4$ & $12$ & $2$ & $8.9$ \\
     & $1.5$B & $6.7$ & $45$ & $4.9$ & $31.7$ \\
     & $3$B & $12.9$ & $90.8$ & $8.2$ & $61.8$ \\
     & $7$B & $28.5$ & $440^{\mathrm{F}}$ & $17.8$ & --- \\
     & $14$B & $55.4$ & --- & $33.2$ & --- \\
     & $32$B & $122.2$ & --- & $71.9$ & --- \\
     & $72$B & $324^{\mathrm{F}}$ & --- & --- & --- \\
     & $110$B & $476^{\mathrm{F}}$ & --- & --- & --- \\
    \midrule
    \multirow{8}{*}{EK-FAC} & $0.5$B & $2.4$ & $12.7$ & $2.1$ & $10.8$ \\
     & $1.5$B & $6.7$ & $45.1$ & $4.9$ & $37.6$ \\
     & $3$B & $12.9$ & $90.8$ & $8.2$ & $73.7$ \\
     & $7$B & $28.5$ & $440^{\mathrm{F}}$ & $17.8$ & --- \\
     & $14$B & $55.4$ & --- & $33.2$ & --- \\
     & $32$B & $122.2$ & --- & $69.8$ & --- \\
     & $72$B & $324^{\mathrm{F}}$ & --- & --- & --- \\
     & $110$B & $476^{\mathrm{F}}$ & --- & --- & --- \\
    \bottomrule
  \end{tabular}
\end{table}

\subsection{Fidelity Evaluation Details}
\label{app:fidelity}

We evaluate three models from three families: GPT-2 ($124$M), Qwen2.5-$1.5$B, and OLMo-3-$7$B~\citep{groeneveld2024olmo}. For each model, all parameters are trained for one epoch on $512$ blocks of $128$ WikiText-2 tokens, drawn with the model's own tokenizer, using batch size $32$ for $16$ AdamW updates. The optimizer uses $\beta_1=0.9$, $\beta_2=0.999$, $\epsilon=10^{-8}$, and zero weight decay. The learning rate peaks at $10^{-5}$ after linear warm-up over the first $10\%$ of steps, followed by linear decay. GPT-2 and Qwen2.5 are trained by the evaluation's own training loop. The OLMo models are trained by OLMo-core, the trainer of the OLMo releases, without modification: the released weights are converted once into an OLMo-core checkpoint, the learning-rate schedule is passed as a scheduler configuration, and \dattri{} captures gradients by wrapping the trainer's \texttt{fit()} call, on one GPU. For TSLOO~\citep{wang2025capturing}, each of $50$ randomly selected training blocks is removed from its original batch in a separate replay of training. We record the change in loss on $64$ held-out validation blocks. Fidelity is the Spearman correlation between attribution scores and these loss changes over the $50$ training blocks, computed separately for each validation block and averaged over the $64$ blocks. For numerical precision, we run training and evaluation using fp32, with TF32 disabled, eager attention, deterministic kernels, and dropout disabled. Pretrained weights are loaded once, and the reference run and all $50$ removal runs start from the same copy. Each model scale uses a fixed seed, and we average over the $64$ query correlations to get robust results.

\paragraph{Methods.}
\dattri{} implements AdamW-influence with full-dimensional gradients and with random coordinate masks. The reduced configuration uses ten disjoint masks of $512$ or $8192$ coordinates per layer, computes scores independently for each mask, and averages them. Its measured cost includes all ten masks. We also evaluate full-dimensional EK-FAC and EK-FAC with factor projection $k_a=k_g=64$. From Bergson (v$0.26.1$), we evaluate MAGIC~\citep{ilyas2025magic}, SOURCE~\citep{bae2024source}, TrackStar~\citep{chang2024trackstar}, and EK-FAC~\citep{george2018fast}. All EK-FAC implementations use damping equal to $0.1$ times the mean eigenvalue of each layer. TrackStar uses per-module projection dimension $64$. SOURCE uses four segments of four steps with two checkpoints per segment. MAGIC uses Bergson's functional trainer to replay the trajectory and differentiate each query loss through the training run; its replay agrees with the reference within $10^{-5}$ in validation loss. On OLMo-3-$7$B the Bergson method that scores a trained model (TrackStar) takes it through its HuggingFace export; MAGIC and SOURCE own the training loop and cannot run under OLMo-core's trainer. Every method scores the same $64$ queries against the same training trajectory.

\paragraph{Batch sizes.}
Every method captures the training gradients in batches of $8$ blocks. The full-dimension methods score the $64$ queries in chunks, and the chunk is, for each library, the largest power of two that completes within the budget: $64$, $8$, and $4$ queries for \dattri{}'s EK-FAC on GPT-2, Qwen2.5-$1.5$B, and OLMo-3-$7$B, and $16$, $1$, and $N/A$ for Bergson's EK-FAC and SOURCE. At $7$B, \dattri{}'s EK-FAC keeps its Kronecker factors in host memory and moves one layer's factors to the GPU at a time.

\paragraph{Resource budget.}
Each execution receives one B200 ($183$\,GB), $16$ CPU cores, $256$\,GB of host memory, and $1$\,TB of local disk; the $7$B model's fp32 training state ($16$ bytes per parameter for weights, gradients, and the two Adam moments) does not fit a smaller GPU. The reported time is the attribution of all $64$ queries: training the trajectory, loading the model and the data, and the TSLOO reference construction are excluded, since a model is trained whether or not it is attributed. For the single-checkpoint methods the time runs from the trained model to the scores; for the trajectory-based methods it covers the attribution pass over the recorded trajectory, which for MAGIC is the backward pass through training. Throughput in Figure~\ref{fig:scaling-fidelity} is $64$ queries divided by this time. Table~\ref{tab:fidelity-limits} further records the limiting resource.

\begin{table}[t]
  \centering\small
  \setlength{\tabcolsep}{5pt}
  \caption{Resource limits in the fidelity benchmark under the budget of one B200, $256$\,GB of host memory, and $1$\,TB of local disk.}
  \label{tab:fidelity-limits}
  \begin{tabular}{lp{2.9cm}p{4.0cm}}
    \toprule
    Method & Model & Limiting resource \\
    \midrule
    MAGIC (Bergson) & Qwen2.5-$1.5$B & GPU memory \\
    AdamW-influence (full, \dattri{}) & Qwen2.5-$1.5$B & GPU memory \\
    MAGIC & OLMo-3-$7$B & GPU memory \\
    AdamW-influence (full) & OLMo-3-$7$B & GPU memory \\
    SOURCE (Bergson) & OLMo-3-$7$B & Local disk \\
    EK-FAC (Bergson) & OLMo-3-$7$B & GPU memory \\
    \bottomrule
  \end{tabular}
\end{table}

\begin{table}[t]
  \centering\small
  \setlength{\tabcolsep}{5pt}
  \caption{Complete fidelity results on one B200: Spearman correlation with TSLOO ($\rho$), attribution time of the $64$ queries in seconds ($t_{\mathrm{attr}}$), and peak GPU memory in GB. ``--'': infeasible under the budget.}
  \label{tab:fidelity-all}
  \begin{tabular}{ll rrr rrr rrr}
    \toprule
    & & \multicolumn{3}{c}{GPT-2} & \multicolumn{3}{c}{Qwen2.5-$1.5$B} & \multicolumn{3}{c}{OLMo-3-$7$B} \\
    \cmidrule(lr){3-5}\cmidrule(lr){6-8}\cmidrule(lr){9-11}
    Library & Method & $\rho$ & $t_{\mathrm{attr}}$ & GB & $\rho$ & $t_{\mathrm{attr}}$ & GB & $\rho$ & $t_{\mathrm{attr}}$ & GB \\
    \midrule
    \dattri{} & AdamW-inf.\ ($k{=}512$)  & .680 & 10 & 8 & .526 & 51 & 47 & .437 & 73 & 136 \\
    \dattri{} & AdamW-inf.\ ($k{=}8192$) & .762 & 28 & 8 & .639 & 103 & 49 & .633 & 170 & 138 \\
    \dattri{} & AdamW-inf.\ (full)       & .845 & 1137 & 25 & -- & -- & -- & -- & -- & -- \\
    \dattri{} & EK-FAC ($k_a{=}k_g{=}64$)        & .699 & 15 & 8 & .420 & 50 & 47 & .303 & 126 & 40 \\
    \dattri{} & EK-FAC (full)            & .664 & 17 & 30 & .434 & 447 & 96 & .402 & 3444 & 139 \\
    \midrule
    Bergson & MAGIC     & .833 & 1265 & 24 & -- & -- & -- & -- & -- & -- \\
    Bergson & SOURCE    & .773 & 846 & 13 & .275 & 28026 & 64 & -- & -- & -- \\
    Bergson & TrackStar & .523 & 34 & 8 & .332 & 289 & 47 & .374 & 368 & 56 \\
    Bergson & EK-FAC    & .681 & 171 & 13 & .432 & 4560 & 64 & -- & -- & -- \\
    \bottomrule
  \end{tabular}
\end{table}

\subsection{Fixed-Batch Comparisons at a Fixed Model Size}
\label{app:fixedbatch}

In addition to the maximum-throughput analysis in the main text, we further evaluate all libraries at the same batch size to examine performance differences under matched conditions. We compare \dattri{}, LogIX, Bergson, and Kronfluence on Pythia-$410$M~\citep{biderman2023pythia}, attributing language-modeling loss on $1024$ WikiText-103 training sequences of length $512$, with either one or $16$ queries. Completed configurations use training batch size $8$ on one A40 ($46$\,GB) in fp32. We evaluate both factor projection with $k_a=k_g=64$ and full-dimensional gradients; projection uses a capture-time Rademacher map where configurable. Table~\ref{tab:crosslib} reports runtime and memory usage for both query counts. Note that these supplementary experiments use a different model and hardware configuration from the Qwen throughput benchmark.

\begin{table}[t]
  \centering
  \small
  \setlength{\tabcolsep}{4.5pt}
  \caption{\textbf{Cross-library efficiency under matched batch sizes.} Results use Pythia-$410$M on one A40 ($46$\,GB), with $1024$ WikiText-103 training sequences and either one or $16$ queries. $t_{\mathrm{attr}}$: attribution wall-clock time (s), excluding model construction and data loading; $M_{\mathrm{GPU}}$: peak GPU memory (GB) over the same phases. The lowest runtime in each row is bold. Completed configurations use training batch size $8$. \oom{} indicates an out-of-memory failure at every tested batch size down to one; \textsc{n/a} indicates an unsupported combination.}
  \label{tab:crosslib}
  \begin{tabular}{llrrrrrrrr}
    \toprule
     & & \multicolumn{2}{c}{\dattri{}}
       & \multicolumn{2}{c}{\textsc{LogIX}}
       & \multicolumn{2}{c}{\textsc{Bergson}}
       & \multicolumn{2}{c}{\textsc{Kronfluence}} \\
    \cmidrule(lr){3-4}\cmidrule(lr){5-6}
    \cmidrule(lr){7-8}\cmidrule(lr){9-10}
    Method & Proj. & $t_{\mathrm{attr}}$ & $M_{\mathrm{GPU}}$ & $t_{\mathrm{attr}}$ & $M_{\mathrm{GPU}}$ & $t_{\mathrm{attr}}$ & $M_{\mathrm{GPU}}$ & $t_{\mathrm{attr}}$ & $M_{\mathrm{GPU}}$ \\
    \midrule
    \multicolumn{10}{c}{\textbf{One query} ($n_{\text{test}}=1$)} \\
    \midrule
    \multirow{2}{*}{\textsc{GradDot}}
      & $k_a{=}k_g{=}64$ & $41$ & $7$ & $68$ & $7$ & $\mathbf{39}$ & $8$
                 & \textsc{n/a} & \textsc{n/a} \\
      & full     & $\mathbf{45}$ & $11$ & \oom{} & --- & $53$ & $13$
                 & $50$ & $11$ \\
    \addlinespace
    \multirow{2}{*}{K-FAC}
      & $k_a{=}k_g{=}64$ & $\mathbf{54}$ & $10$ & $89$ & $7$ & $180$ & $20$
                 & \textsc{n/a} & \textsc{n/a} \\
      & full     & $\mathbf{134}$ & $19$ & \oom{} & --- & $194$ & $20$
                 & $201$ & $12$ \\
    \addlinespace
    \multirow{2}{*}{EK-FAC}
      & $k_a{=}k_g{=}64$ & $\mathbf{58}$ & $10$ & $143$ & $7$
                 & \textsc{n/a} & \textsc{n/a}
                 & \textsc{n/a} & \textsc{n/a} \\
      & full     & $\mathbf{223}$ & $21$ & \oom{} & --- & $278$ & $27$
                 & $347$ & $15$ \\
    \midrule
    \multicolumn{10}{c}{\textbf{16 queries} ($n_{\text{test}}=16$)} \\
    \midrule
    \multirow{2}{*}{\textsc{GradDot}}
      & $k_a{=}k_g{=}64$ & $40$ & $13$ & $71$ & $7$ & $\mathbf{39}$ & $8$
                 & \textsc{n/a} & \textsc{n/a} \\
      & full     & $58$ & $31$ & \oom{} & --- & $121$ & $30$
                 & $\mathbf{57}$ & $28$ \\
    \addlinespace
    \multirow{2}{*}{K-FAC}
      & $k_a{=}k_g{=}64$ & $\mathbf{55}$ & $18$ & $90$ & $7$ & $225$ & $44$
                 & \textsc{n/a} & \textsc{n/a} \\
      & full     & $\mathbf{153}$ & $34$ & \oom{} & --- & $317$ & $44$
                 & $210$ & $28$ \\
    \addlinespace
    \multirow{2}{*}{EK-FAC}
      & $k_a{=}k_g{=}64$ & $\mathbf{59}$ & $18$ & $146$ & $7$
                 & \textsc{n/a} & \textsc{n/a}
                 & \textsc{n/a} & \textsc{n/a} \\
      & full     & $\mathbf{230}$ & $38$ & \oom{} & --- & $461$ & $44$
                 & $360$ & $28$ \\
    \bottomrule
  \end{tabular}
\end{table}

\paragraph{Results.}
\dattri{} has the lowest runtime in nine of the twelve rows of Table~\ref{tab:crosslib}, and in every row that uses a curvature-based method. For projected K-FAC and EK-FAC it runs $1.6$--$2.5\times$ faster than LogIX and $3.3$--$4.1\times$ faster than Bergson; for the full-dimensional variants it runs $1.4$--$1.6\times$ faster than Kronfluence and $1.2$--$2.1\times$ faster than Bergson. The remaining three rows are \textsc{GradDot}, where the runtimes of \dattri{}, Bergson, and Kronfluence differ by at most two seconds. At this scale a single forward and backward pass over the $1024$ training sequences in fp32 accounts for most of the runtime, and \textsc{GradDot} adds little to it, so any library that runs the model once lands near the same floor; LogIX, which extracts gradients one query at a time through a disk-backed store, does not. The curvature-based methods add a fit over the training gradients and a per-layer preconditioning of the queries, and it is this attribution-side work that separates the libraries. Moving from one query to $16$ leaves the runtime of \dattri{} almost unchanged, whereas Bergson's full-dimensional K-FAC and EK-FAC take $1.6$--$1.7\times$ longer.

\subsection{Routing Across Sequence Lengths}
\label{app:routingexp}

\begin{figure}
    \centering
    \includegraphics[width=\linewidth]{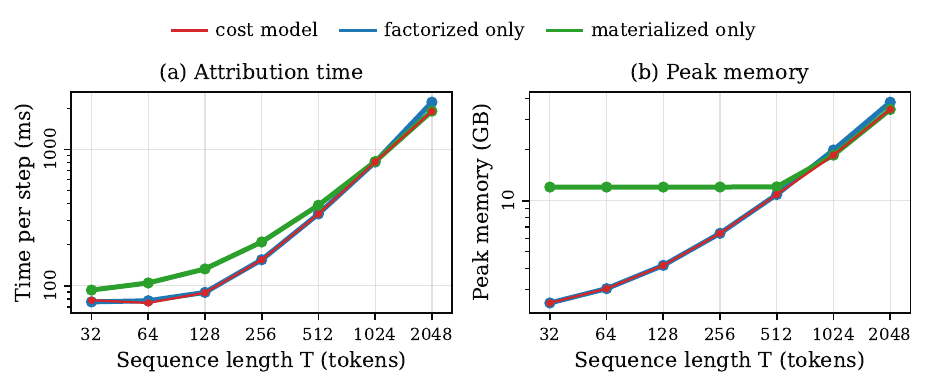}
    \caption{\textbf{Adaptive routing versus fixed gradient representations.} Mean attribution time per step (left) and peak GPU memory (right) for full-dimensional \textsc{GradDot} on Pythia-$410$M as sequence length increases, using batch size $8$, one query, and fp32 on one A40. The cost-model heuristic selects between factorized and materialized execution; fixed-route ablations use the same implementation. Adaptive routing closely tracks the faster fixed configuration while maintaining the lowest or tied peak memory at every evaluated length. Memory includes all GPU allocations.}
    \label{fig:routing}
\end{figure}

In this section, we show the effectiveness of the heuristic proposed in \S\ref{sec:efficiency}. We use the full-dimensional \textsc{GradDot} setting of Appendix~\ref{app:fixedbatch}: Pythia-$410$M, WikiText-103, fp32, one query, and one A40. Sequence length ranges from $32$ to $2048$ tokens, and all configurations use batch size $8$. We time $128$ steps after $8$ warm-up steps and report mean attribution time per step and peak GPU memory. The three \dattri{} configurations differ only in their routing decision; the fixed-route runs are ablations of the same implementation. Figure~\ref{fig:routing} presents the curves.

\paragraph{Adaptive versus fixed execution.}
The maximum runtime excess over the faster fixed configuration is approximately $2.8\%$. For $32\le T\le512$, materialized execution takes $16$--$50\%$ longer than adaptive routing. At $T=2048$, factorized execution takes approximately $17\%$ longer. Noticeably, while the heuristic is developed for runtime optimization, the short-sequence memory advantage is also substantial: at $T=32$, the always-materialized run has $4.8\times$ the peak memory of adaptive execution. These are total GPU-memory measurements, not isolated gradient-buffer sizes. Adaptive routing has the lowest or tied memory among the three \dattri{} configurations at every evaluated length.

\subsection{Token-Level Attribution Examples}
\label{app:tokenmore}

We illustrate token-level attribution using \textsc{GradDot}, K-FAC, and EK-FAC on Pythia-$410$M and WikiText-103. For these methods, the sequence-level score decomposes over the training sequence's token positions. For a single layer, let $R^{(j)}\in\mathbb{R}^{N_o\times N_i}$ denote the query-side matrix for query $j$. The contribution of position $t$ in training sample $i$ is
\[
    s_t^{(i,j)}=(g_t^{(i)})^\top R^{(j)}a_t^{(i)},
    \qquad
    \sum_{t=1}^{T_i}s_t^{(i,j)}
    =\left\langle\mathrm{D}W^{(i)},R^{(j)}\right\rangle_F.
\]
Here $R^{(j)}=\mathrm{D}W^{(j)}$ for \textsc{GradDot}; for K-FAC and EK-FAC, it is the corresponding preconditioned query gradient reshaped into a matrix. Contributions are summed across attributed layers to obtain the reported token scores, whose sum recovers the sequence-level score. All attention and MLP linear layers are attributed without projection. K-FAC and EK-FAC fit curvature on $32$ WikiText-103 paragraphs with damping $10^{-3}$. Each panel is normalized independently because the methods produce scores on different scales. The figures illustrate reuse of the captured factors for fine-grained analysis; they do not establish a token decomposition for every attribution method in the library.

\begin{figure}[t]
  \centering
  \includegraphics[width=\textwidth]{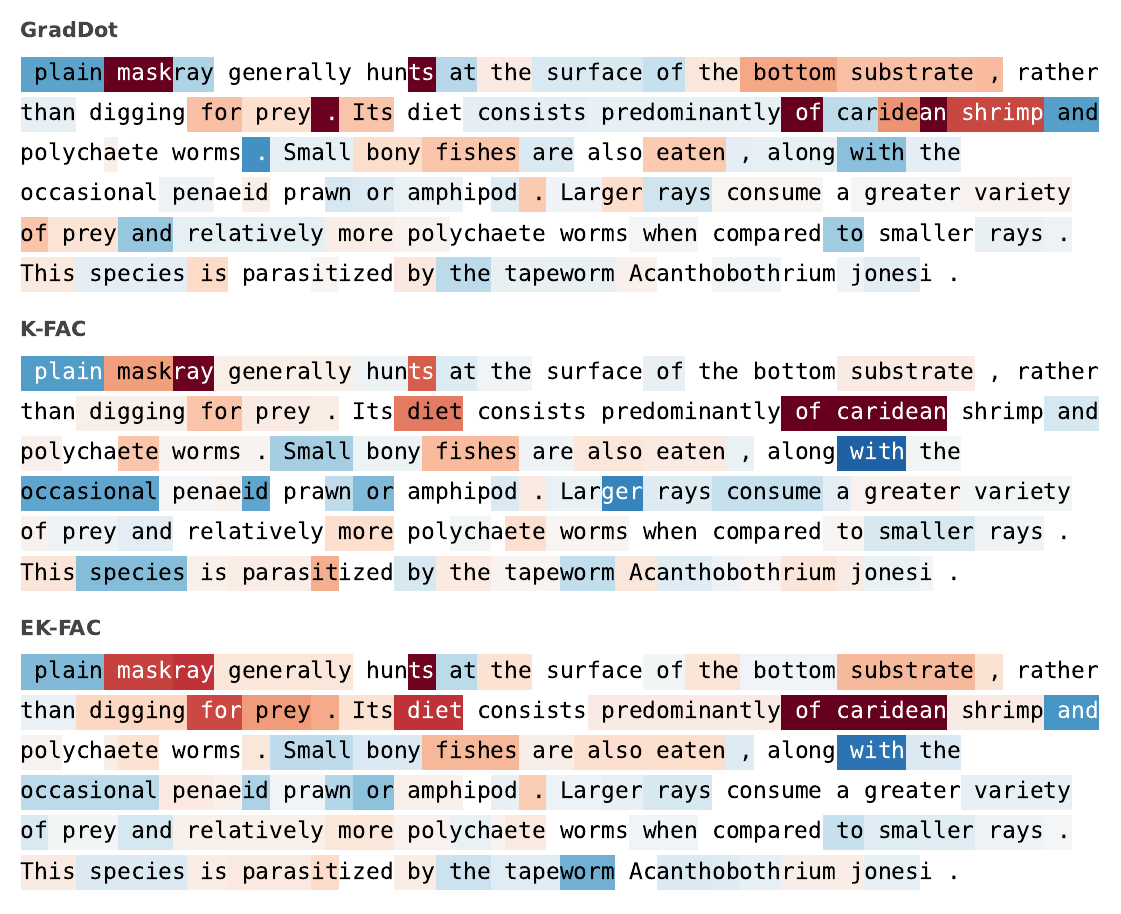}
  \caption{\textbf{maskray} (train row 308). Target: ``The plain maskray is a species of stingray that feeds on caridean $\to$ \textbf{shrimp}''. Under \textsc{GradDot} the largest positions are the subword \code{ts} of \emph{hunts}, \emph{mask}, and the two occurrences of \emph{of}, with \emph{shrimp} itself at rank $6$. Both preconditioned methods concentrate on \emph{of caridean}, the phrase the target completes.}
  \label{fig:tok-maskray}
\end{figure}

\begin{figure}[t]
  \centering
  \includegraphics[width=\textwidth]{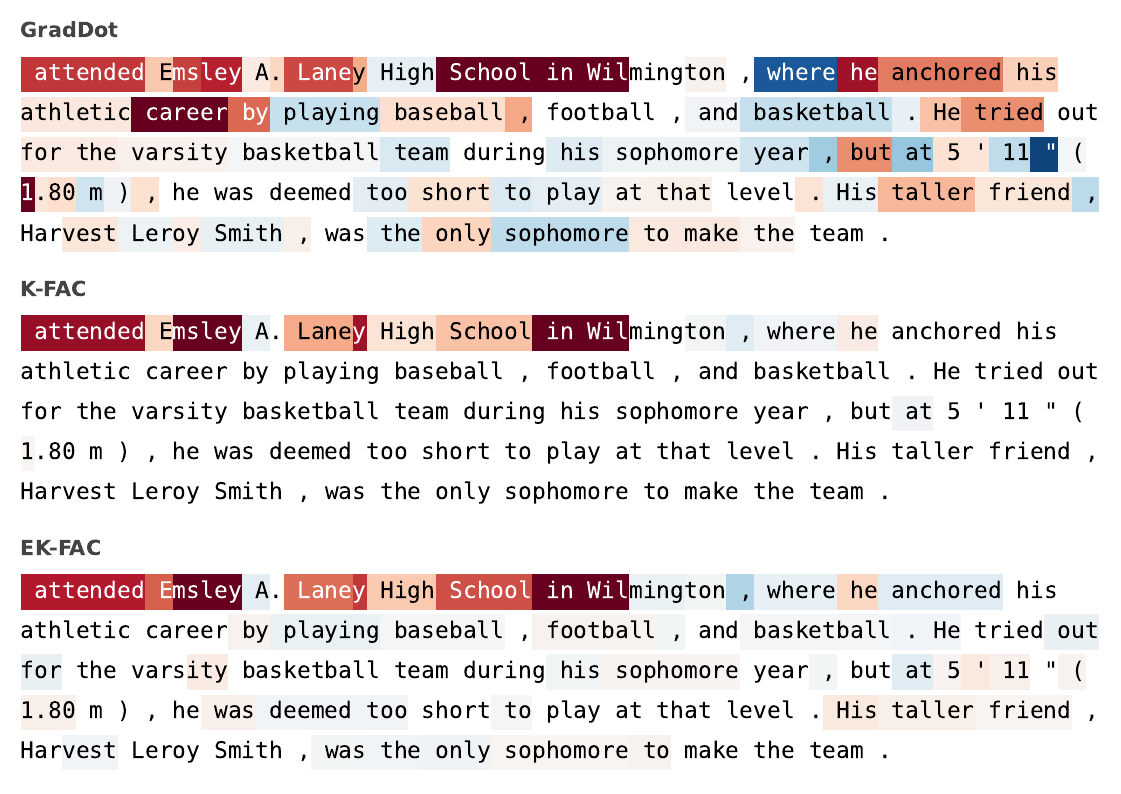}
  \caption{\textbf{jordan} (train row 1724). Target: ``Michael Jordan attended Emsley A.\ Laney High School in $\to$ \textbf{Wilmington}''. The paragraph states the fact. \textsc{GradDot} ranks \code{Wil} second, behind the function word \emph{in}, and shades the whole paragraph; K-FAC and EK-FAC put \code{Wil} first, followed by \emph{in} and the remaining subwords of the city name, and leave the text after the school name essentially blank.}
  \label{fig:tok-jordan}
\end{figure}

\begin{figure}[t]
  \centering
  \includegraphics[width=\textwidth]{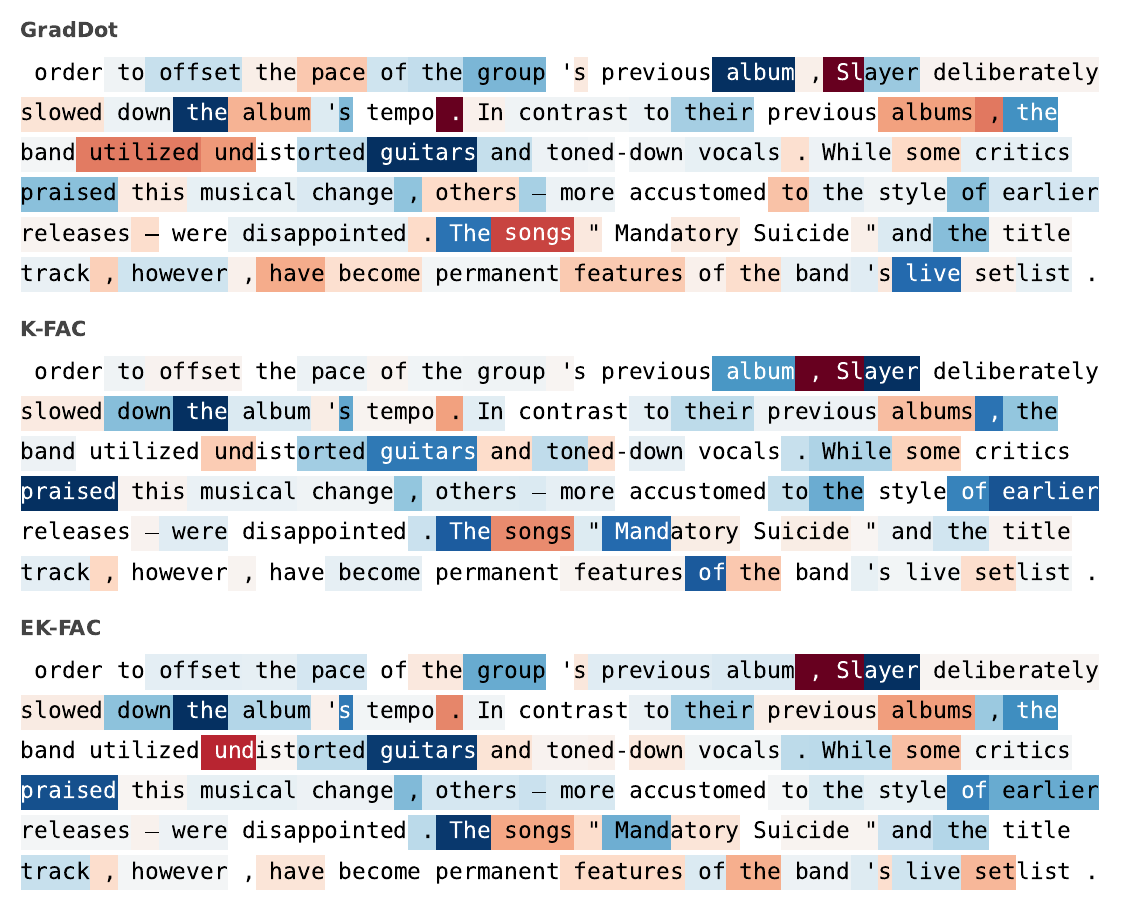}
  \caption{\textbf{slayer} (train row 752). Target: ``South of Heaven is a studio album by the American thrash metal band $\to$ \textbf{Slayer}'' (tokenized \code{Sl}\,\code{ayer}). The answer's first subword ranks first under all three attributors, but its lead over the runner-up grows from $3.6\times$ (\textsc{GradDot}) to $7.3\times$ (EK-FAC) and $35.7\times$ (K-FAC). The paragraph discusses the album's tempo rather than restating the band's name, and the sequence-level \textsc{GradDot} score is in fact negative ($-82$), whereas both preconditioned scores are positive: preconditioning concentrates positive attribution on the answer token, while the unpreconditioned sequence-level score is negative.}
  \label{fig:tok-slayer}
\end{figure}

\subsection{Weight-Gradient-Free Capture for Frozen Attribution}
\label{app:capture-path}

Attribution at a fixed checkpoint reads only the inputs and output gradients of the hooked layers; the weight gradients that the backward pass also computes are never used. \dattri{} therefore provides an optional capture mode (\texttt{invasive\_linear\_io}) that, during capture, routes each hooked \texttt{nn.Linear} through an operation whose backward pass returns only the input gradient. It enables more efficient exact attribution by eliminating the memory required for parameter gradients and reducing the runtime overhead of gradient collection. Because the mode produces no parameter gradients, it is applicable only outside training; capture within a training loop, including online data selection, uses standard capture. Kronfluence and Bergson avoid the same computation by freezing the model parameters during attribution, whereas LogIX has to compute weight gradients for the layers it tracks. In Table~\ref{tab:crosslib}, \dattri{} uses this mode for \textsc{GradDot} and full-dimensional K-FAC and EK-FAC, and standard capture for projected K-FAC and EK-FAC. To quantify its effect, we rerun each \dattri{} configuration of Table~\ref{tab:crosslib} under both modes, changing only the capture mode, on one L40S (48\,GB) in fp32 with five paired repetitions. As Table~\ref{tab:capture-speedup} shows, standard capture takes 7--19\% longer and uses at most 4.5\,GB more peak memory, which is the cost of the weight-gradient computation it retains.

\begin{table}[t]
  \centering\small
  \setlength{\tabcolsep}{5pt}
  \caption{Speedup of weight-gradient-free capture over standard capture (standard time / weight-gradient-free time) for \dattri{} on Pythia-$410$M, one L40S (48\,GB), fp32, with the workload of Table~\ref{tab:crosslib}. Each entry is the median over five paired repetitions.}
  \label{tab:capture-speedup}
  \begin{tabular}{llcc}
    \toprule
    Method & Projection & $n_{\text{test}}=1$ & $n_{\text{test}}=16$ \\
    \midrule
    \textsc{GradDot} & $k_a=k_g=64$ & $1.19\times$ & $1.18\times$ \\
    \textsc{GradDot} & full   & $1.16\times$ & $1.09\times$ \\
    K-FAC            & $k_a=k_g=64$ & $1.14\times$ & $1.11\times$ \\
    K-FAC            & full   & $1.10\times$ & $1.07\times$ \\
    EK-FAC           & $k_a=k_g=64$ & $1.10\times$ & $1.14\times$ \\
    EK-FAC           & full   & $1.08\times$ & $1.08\times$ \\
    \bottomrule
  \end{tabular}
\end{table}

\subsection{Data Selection Details}
\label{app:dataselect}

Training uses Llama-3.2-1B with AdamW, a linear learning-rate schedule with $3\%$ warm-up, gradient clipping at $1.0$, and batch size $8$, over three seeds. The learning rate is $10^{-6}$ for full-parameter fine-tuning and $10^{-4}$ for LoRA. Master weights are stored in fp32, with bf16 autocast during training. The source pool is Alpaca and the target task is SAMSum, with the same chat template applied to both datasets. At each step, the scoring target is one example from a $16$-example SAMSum validation pool. Selection removes the bottom half of each batch according to \textsc{GradDot} scores against that example. Full-parameter runs attribute all transformer attention and MLP linear layers, excluding embeddings and the language-model head. LoRA uses rank $8$ and $\alpha=16$ on the same projections; capture and gradient intervention are restricted to the adapter layers. Test perplexity is evaluated every $50$ steps on the response tokens of $500$ held-out SAMSum examples, with prompt tokens masked.

\begin{figure}[t]
  \centering
  \includegraphics[width=0.65\linewidth]{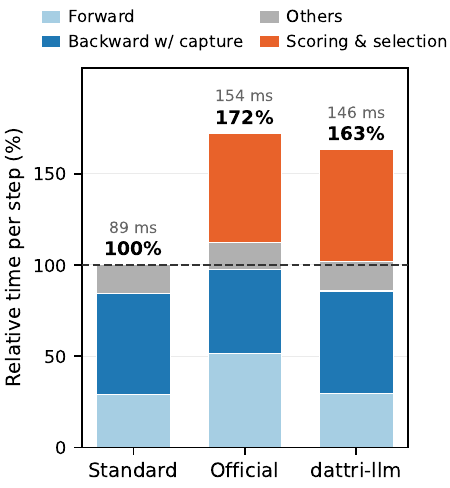}
  \caption{Per-step runtime of online data selection relative to standard training, under full-parameter fine-tuning.}
  \label{fig:runtime}
\end{figure}

\paragraph{Runtime.}
Figure~\ref{fig:runtime} reports per-step runtime under full-parameter fine-tuning, averaged over steps after warm-up. Standard training takes $89$\,ms per step, selection with \dattri{} takes $146$\,ms, and the dedicated implementation of \citet{hu2026dr} takes $154$\,ms. The library implementation is approximately $5\%$ faster than the dedicated implementation. The dedicated implementation uses an additional forward pass to obtain training gradients, whereas \dattri{} captures the required quantities from the training backward pass. Scoring and selection account for the remaining application overhead.

\end{document}